\documentclass[letterpaper]{article} 
\usepackage{aaai2026}  
\usepackage{times}  
\usepackage{helvet}  
\usepackage{courier}  
\usepackage[hyphens]{url}  
\usepackage{graphicx} 
\usepackage{natbib}  
\usepackage{caption} 
\usepackage{newfloat}
\usepackage{listings}

\usepackage{subfigure}
\usepackage{multirow}
\usepackage{booktabs}
\usepackage{pifont}
\usepackage{paracol}
\usepackage{amsmath,amssymb}

\usepackage{pdfpages}

\usepackage{algorithm}
\usepackage{algpseudocode}

\newcommand{\cmark}{\ding{51}}%
\newcommand{\xmark}{\ding{55}}%
\renewcommand{\vec}[1]{\boldsymbol{#1}}

\DeclareCaptionStyle{ruled}{labelfont=normalfont,labelsep=colon,strut=off} 
\floatstyle{ruled}
\newfloat{listing}{tb}{lst}{}
\floatname{listing}{Listing}

\title{Diversifying Counterattacks: Orthogonal Exploration for Robust CLIP Inference}
\author{
    Chengze Jiang\textsuperscript{\rm 1}, Minjing Dong\textsuperscript{\rm 2}, Xinli Shi\textsuperscript{\rm 1}, Jie Gui\equalcontrib\textsuperscript{\rm 1,3,4}\\
}
\affiliations{
    \textsuperscript{\rm 1}  Southeast University, Nanjing, China\\
    \textsuperscript{\rm 2}  Department of Computer Science, City University of Hong Kong, Hong Kong\\
	\textsuperscript{\rm 3}  Engineering Research Center of Blockchain Application, Supervision And Management (Southeast University), Ministry of Education, China\\
	\textsuperscript{\rm 4}  Purple Mountain Laboratories, China \\
	czjiang@seu.edu.cn, minjdong@cityu.edu.hk, xinli\_shi@seu.edu.cn, guijie@seu.edu.cn
}

\usepackage{bibentry}
\begin{document}
\maketitle

\begin{abstract}
Vision-language pre-training models (VLPs) demonstrate strong multimodal understanding and zero-shot generalization, yet remain vulnerable to adversarial examples, raising concerns about their reliability. Recent work, Test-Time Counterattack (TTC), improves robustness by generating perturbations that maximize the embedding deviation of adversarial inputs using PGD, pushing them away from their adversarial representations. However, due to the fundamental difference in optimization objectives between adversarial attacks and counterattacks, generating counterattacks solely based on gradients with respect to the adversarial input confines the search to a narrow space. As a result, the counterattacks could overfit limited adversarial patterns and lack the diversity to fully neutralize a broad range of perturbations. In this work, we argue that enhancing the diversity and coverage of counterattacks is crucial to improving adversarial robustness in test-time defense. Accordingly, we propose Directional Orthogonal Counterattack (DOC), which augments counterattack optimization by incorporating orthogonal gradient directions and momentum-based updates. This design expands the exploration of the counterattack space and increases the diversity of perturbations, which facilitates the discovery of more generalizable counterattacks and ultimately improves the ability to neutralize adversarial perturbations. Meanwhile, we present a directional sensitivity score based on averaged cosine similarity to boost DOC by improving example discrimination and adaptively modulating the counterattack strength. Extensive experiments on 16 datasets demonstrate that DOC improves adversarial robustness under various attacks while maintaining competitive clean accuracy. Code is available at \url{https://github.com/bookman233/DOC}.
\end{abstract}

\section{Introduction}
Vision-language pre-training models (VLPs) have emerged as powerful multimodal systems, demonstrating strong zero-shot generalization \cite{zhang2024vision, yang2025visionzip, laurenccon2024matters}. Among them, CLIP is a representative VLP that aligns visual and textual representations through contrastive learning and achieves impressive performance in vision tasks \cite{CLIPRef, jiao2023learning}. While recent research primarily focuses on improving the performance of CLIP models \cite{zhou2023zegclip}, their adversarial robustness receives comparatively less attention \cite{dongimproving}. Recent studies reveal that CLIP is vulnerable to adversarial examples, \emph{i.e.}, human-imperceptible perturbations that can mislead predictions of the model \cite{yu2024text, zhang2024robust, yang2024prompt}. This vulnerability raises concerns about the reliability of CLIP \cite{li2024language, zhang2025enhancing, ge2023improving}. Since an increasing number of CLIP models are deployed in security-related downstream tasks, enhancing their adversarial robustness has become an urgent research priority \cite{wortsman2022robust}.
\par
One representative solution is adversarial fine-tuning, which improves adversarial robustness by fine-tuning the pretrained CLIP model using adversarial examples \cite{maounderstanding, schlarmann2024robust}. Another approach is adversarial prompt tuning, which introduces learnable text tokens into the embedding space and uses a small validation set to better align prompt embeddings with those of adversarial images \cite{li2024one, sheng2025r}. Although these methods improve the adversarial robustness of CLIP, they still present notable limitations. First, adversarial fine-tuning introduces significant computational overhead, which grows with the size of the dataset \cite{alfarra2022combating, zhangclipure}. In contrast, prompt tuning requires only a few labeled examples to adjust the prompt, thereby reducing the computational cost \cite{wang2025tapt}. However, it operates in the learned embedding space rather than the human-interpretable textual domain, causing the learned prompts to lose semantic interpretability \cite{raman2023model}. Most importantly, although CLIP benefits from large-scale pretraining that gives it impressive generalization ability \cite{CLIPRef, hu2022scaling}, fine-tuning its model weights can diminish this generalization \cite{wang2024vilt}. Recently, Test-Time Counterattack (TTC) is presented as a parameter-free and data-agnostic defense that leverages the expressive power of CLIP to improve adversarial robustness \cite{xing2025clip}. TTC fixes the adversarial input as an anchor and optimizes a counterattack using PGD \cite{madry2018towards} to maximize the $\ell_2$ distance between the adversarial input and its counterattacked variants, thereby pushing adversarial input away from the adversarial neighborhood. 
\par
\begin{figure}[t]\centering
    \subfigure[Conceptual illustration of our DOC]{\includegraphics[scale=0.58]{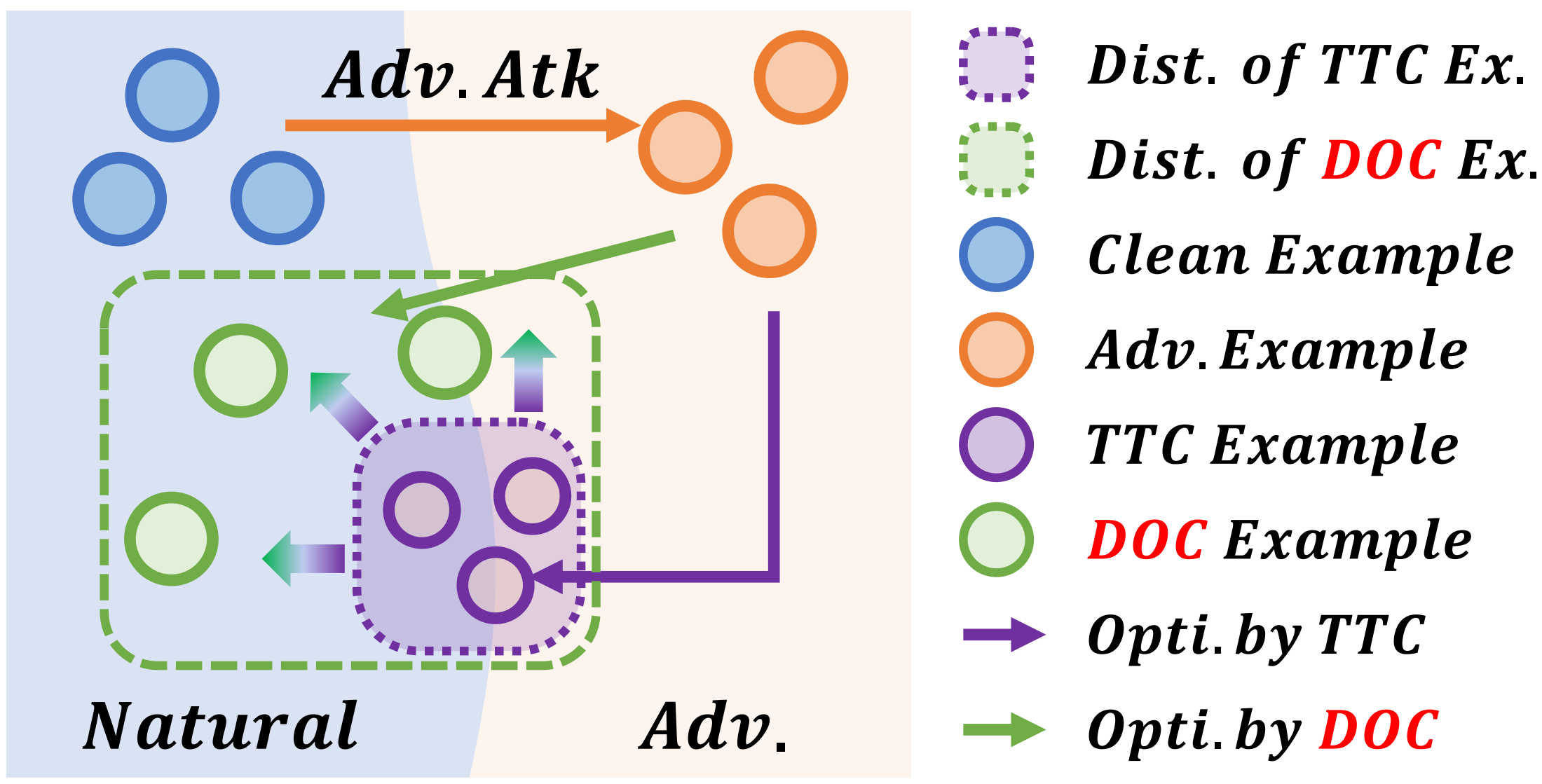}}
    \subfigure[Distributions]{\includegraphics[scale=0.58]{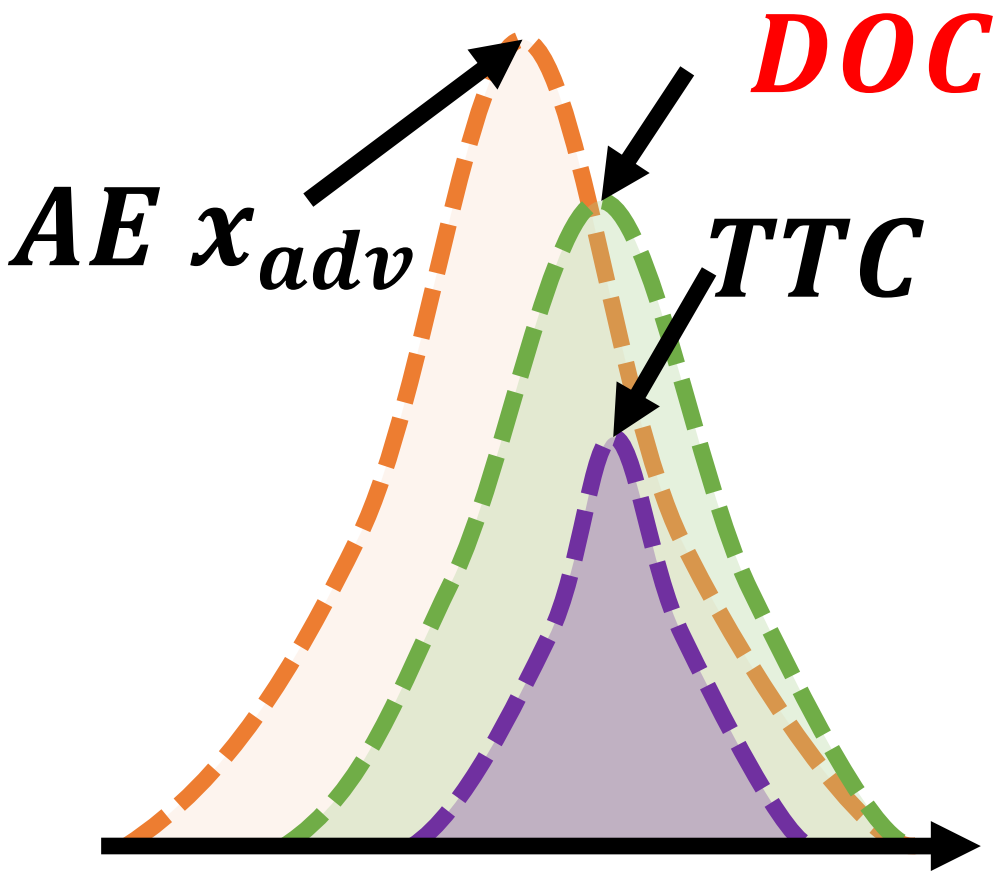}}
    \subfigure[t-SNE on CIFAR10]{\includegraphics[scale=0.27]{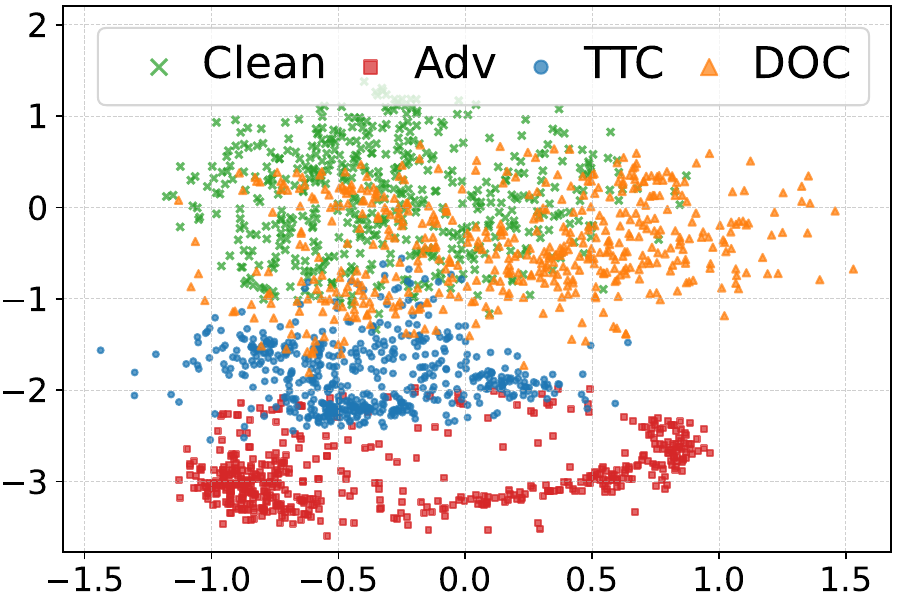}}
    \subfigure[t-SNE on STL10]{\includegraphics[scale=0.27]{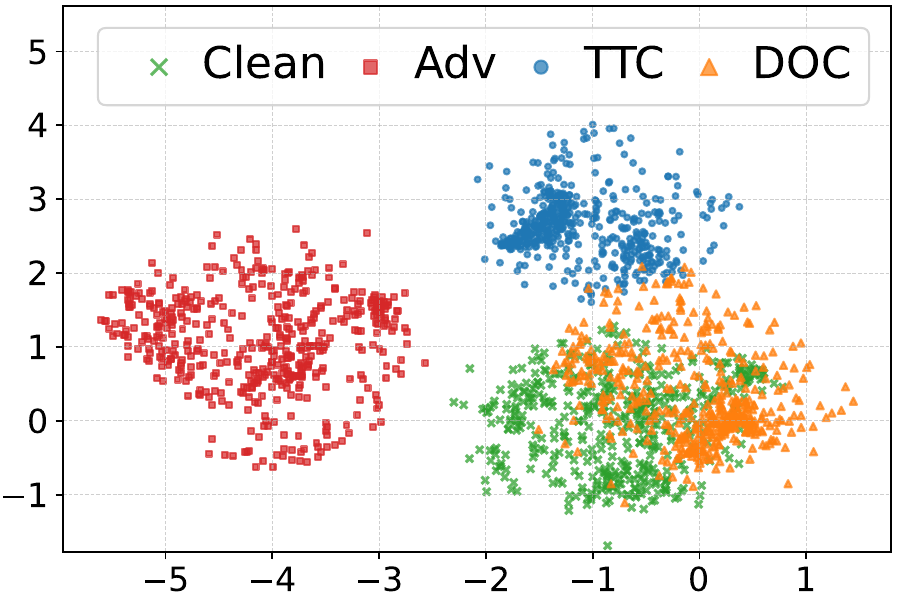}}
    \caption{(a)-(b) Conceptual illustration of our methodology. We propose to generate more diverse counterattacks to neutralize adversarial perturbations. (c)-(d) t-SNE of example embeddings obtained by TTC and our DOC. 
    }
    \label{Motivation}
\end{figure}
While TTC presents promising progress, there exists a fundamental mismatch between the optimization objectives of adversarial attack and counterattack. Specifically, adversarial attacks aim at maximizing the loss (defined in equation \eqref{VLMAtk}), while counterattacks aim at maximizing the distance between adversarial and counterattack examples (defined in equation \eqref{TTCOptimization}). This mismatch could even be further amplified regarding the optimization strategy in TTC since it uses PGD to generate counterattacks and could overfit to the surrogate objective easily, which can hardly approximate the accurate adversarial perturbation distribution. Ultimately, this mismatch hinders the counterattack from effectively neutralizing the underlying adversarial perturbations. Thus, in the absence of label supervision at test time, refining the optimization strategy of counterattacks becomes crucial to alleviate overfitting induced by the mismatch of inherent optimization objectives. A natural and direct approach is to augment the optimization process to increase counterattack diversity, enabling broader exploration of the adversarial perturbation space and enhancing the ability to neutralize a wide range of potential threats (as shown in Fig. \ref{Motivation}(a) and (b)). Therefore, improving counterattack diversity to more effectively defend against adversarial threats of CLIP remains an open and valuable research challenge. 
\par
\begin{figure}[t]\centering
    \subfigure[CIFAR10]{\includegraphics[scale=0.16]{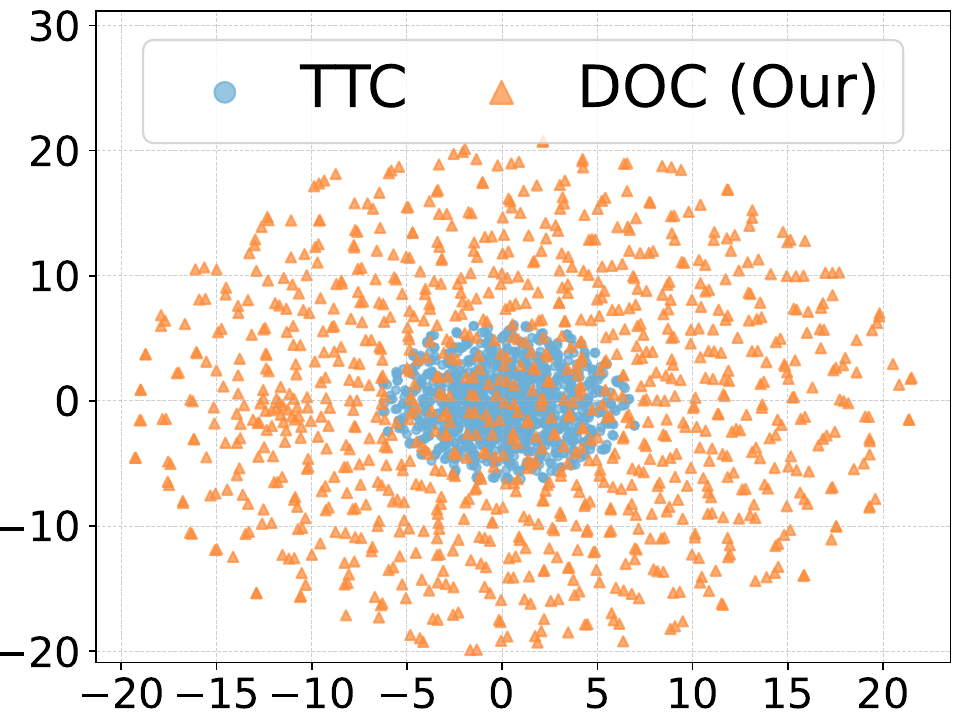}}
    \subfigure[STL10]{\includegraphics[scale=0.16]{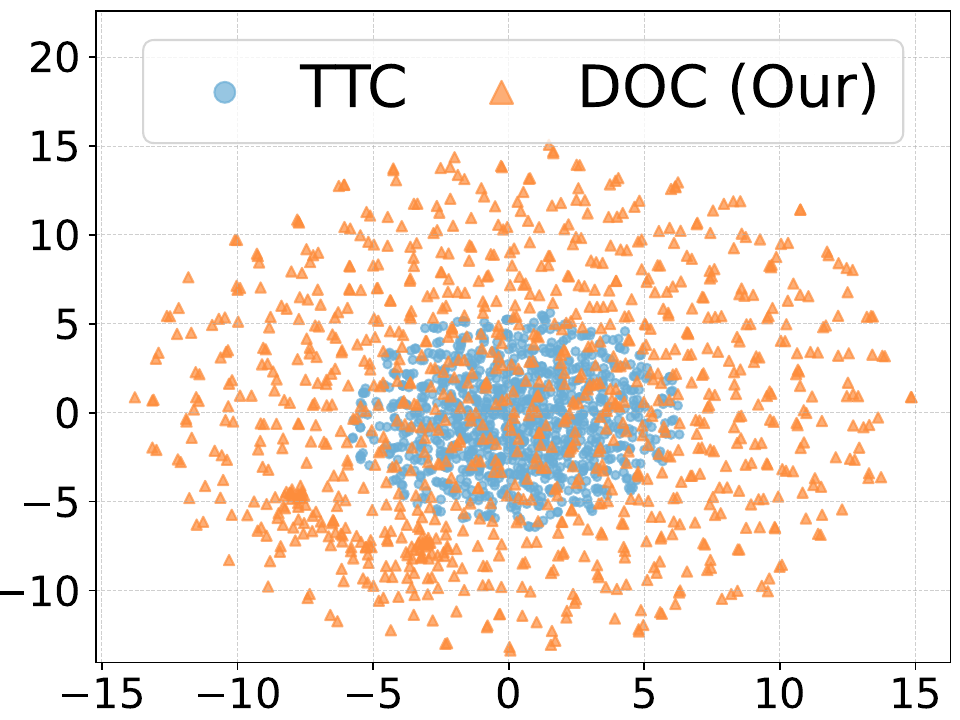}}
    \subfigure[ImageNet]{\includegraphics[scale=0.16]{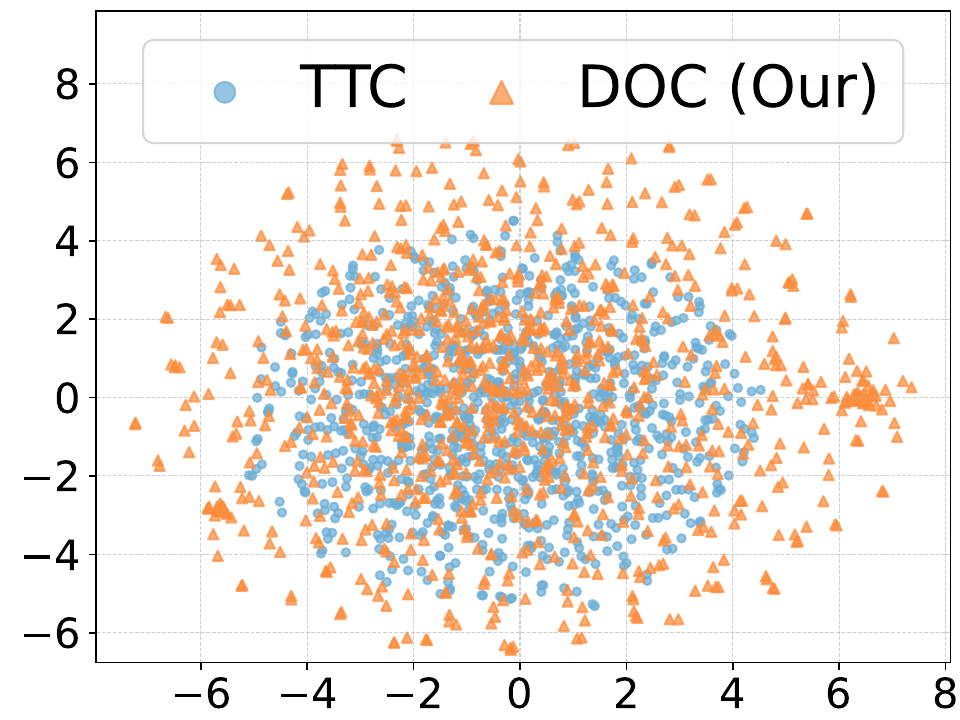}}
    \includegraphics[scale=0.63]{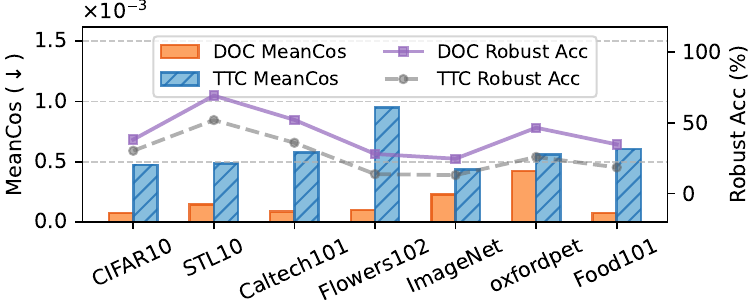}
    \caption{(a)-(c) t-SNE visualizations of counterattacks generated by TTC and our DOC. (Bottom) Comparison of mean cosine similarity of counterattack and robust accuracy under PGD-10 with $\epsilon_{\text{atk}} = 4/255$. More details on 15 datasets are presented in \textbf{Supplementary Materials}.}
    \label{tSNE}
\end{figure}

Consequently, we introduce Directional Orthogonal Counterattack (DOC), which augments each optimization step of counterattack with a randomized component orthogonal to the primary gradient direction and incorporates a momentum-based update. This design expands the counterattack search space to increase distribution diversity, allowing the counterattack to escape narrow local optima and more effectively neutralize adversarial effects in an unsupervised setting (as shown in Fig. \ref{Motivation}). As further illustrated in Fig. \ref{tSNE}, t-SNE visualizations and mean cosine similarity (MeanCos, where lower values indicate higher diversity \cite{MeanCos, zhu2023boosting}) show that DOC generates more diverse counterattacks compared to TTC, resulting in improved adversarial robustness of CLIP. Furthermore, DOC introduces a directional sensitivity score, defined as the cosine similarity between the original image embedding and its randomly perturbed versions, which guides the adaptive modulation of counterattack strength. Comprehensive evaluations on 16 datasets confirm that the components of DOC jointly improve the test-time robustness of CLIP models while preserving competitive clean accuracy. The main contributions are summarized as follows:
\begin{itemize}
   \item We propose DOC to more effectively neutralize adversarial perturbations by expanding the counterattack search space and increasing diversity through the incorporation of orthogonal components and momentum.
    \item We introduce the directional sensitivity score via cosine similarity, which determines the necessity of a counterattack and enables fine-grained control over its strength.
    \item Experiments on 16 datasets show that DOC outperforms state-of-the-art test-time defenses in adversarial robustness while maintaining competitive clean accuracy. 
\end{itemize}


\section{Related Works}
\subsection{Adversarial Robustness}
Deep neural networks are vulnerable to adversarial attacks \cite{cui2024robustness, jiang2025improving, xia2024transferable}. To mitigate this vulnerability, adversarial training is recognized as one of the most effective defenses \cite{tong2024taxonomy, xhonneux2024efficient, kuang2024defense}. However, it imposes significant computational costs and often struggles with overfitting \cite{wang2024pre, jia2024improving}. In parallel, test-time defenses have attracted increasing attention because they do not require modifying model parameters \cite{croce2022evaluating}, including adversarial purification \cite{nie2022diffusion} and loss-based adjustment \cite{wu2021attacking, alfarra2022combating}. Despite their progress, existing test-time defenses remain susceptible to attacks designed to circumvent their mechanisms. For example, Hedge Defense (HD) optimizes test-time perturbations by maximizing the loss across all classes \cite{wu2021attacking}. While promising, HD relies on classification-oriented objectives and assumes access to supervised information or adversarially trained backbones. Although adversarial defense methods have made progress, most existing approaches focus on unimodal supervised settings and face challenges when generalizing to modern vision-language models, which rely on multimodal embedding architectures and do not depend on supervised information for inference.

\subsection{Adversarial Robustness of VLPs}
VLPs demonstrate strong zero-shot generalization capabilities \cite{zhang2024long, yang2024clip} but remain vulnerable to adversarial attacks \cite{tu2023closer, zhang2025enhancing}. Therefore, various defense strategies are presented to improve the robustness of VLPs. Among them, adversarial fine-tuning trains the model with adversarial examples to strengthen robustness \cite{maounderstanding, gong2025boosting}. TeCoA demonstrates transferability across tasks \cite{maounderstanding}, and PMG-AFT adds CLIP-guided regularization to relieve overfitting \cite{wang2024pre}. Another approach is adversarial prompt tuning \cite{zhang2024adversarial}, which adjusts input prompts and learns optimized prompt tokens to better align text and image features under adversarial conditions \cite{wang2025tapt, sheng2025r}. Despite these advances, existing methods require supervised training, access to downstream tasks, or rely on prompt engineering, which risks undermining the generalization of models or introducing additional training processes \cite{mou2024sg}. To address this limitation, recent work by Liu $et~al.$ introduces TTC, which neutralizes adversarial perturbations by counterattack, achieving defense without changing model parameters or using prompt engineering \cite{xing2025clip}. However, a challenge is that the distributional shift between adversarial and clean examples makes using the adversarial embedding as an anchor risk overfitting to the local adversarial structure. Motivated by this, we aim to enhance counterattack diversity to broaden the search space and improve the neutralization of adversarial noise, thereby boosting CLIP’s adversarial robustness.

\section{Methodology}
\subsection{Background and Preliminaries}
\subsubsection{Background}
CLIP is a representative foundation VLP that achieves impressive zero-shot performance through large-scale pretraining on paired image-text data \cite{cao2024empirical}, which comprises an image encoder $I_\theta: \mathcal{X} \to \mathbb{R}^d$ and a text encoder $T_\phi: \mathcal{T} \to \mathbb{R}^d$, parameterized by $\theta$ and $\phi$, respectively \cite{gao2024clip}. For inference, given an input image $\vec{x} \in \mathcal{X}$ and a textual prompt $\vec{t}_i \in \mathcal{T}$ representing the $i$-th class, CLIP computes their cosine similarity as follows:
\begin{equation}
    s(\vec{x}, \vec{t}_i) = \frac{\langle I_\theta(\vec{x}), T_\phi(\vec{t}_i) \rangle}{\|I_\theta(\vec{x})\| \cdot \|T_\phi(\vec{t}_i)\|},
\end{equation}
where $\vec{t}_i$ denotes the textual prompt for the $i$-th class \cite{CLIPRef}. The similarity across all candidate classes is normalized to yield the predicted class distribution as $P(y = i \mid \vec{x}) = \exp(s)/\sum_j \exp(s)$. The predicted label is determined as the class with the highest probability. 

\subsubsection{Adversarial Vulnerability of VLPs}
To evaluate the adversarial robustness of VLPs, an adversary obtains adversarial perturbation $\vec{\delta}_{\text{adv}}$, bounded by an $\ell_p$ norm, such that the adversarial example $\vec{x}_{\text{adv}} = \vec{x} + \vec{\delta}_{\text{adv}}$ leads to incorrect predictions \cite{gao2024boosting, guo2024efficient}. The objective of an adversarial attack is typically formulated as the following constrained maximization problem \cite{zhao2023evaluating}:
\begin{equation}\label{VLMAtk}
    \vec{\delta}_{\text{atk}} = \arg\max_{\vec{\delta}} \mathcal{L}(\vec{x} + \vec{\delta}, y), \quad \text{s.t.} \quad \|\vec{\delta}_{\text{atk}}\|_p \leq \epsilon_{\text{atk}},
\end{equation}
where $y$ denotes the label, $\mathcal{L}$ is the loss function, and $\epsilon_{\text{atk}}$ is the adversarial perturbation budget \cite{wang2024transferable}. By optimizing the objective \eqref{VLMAtk}, various adversarial attacks can generate perturbations $\vec{\delta}_{\text{adv}}$ that are injected into the original input to create adversarial examples that mislead the VLPs.

\subsubsection{Test-Time Counterattacks for CLIP}
Recently, TTC is presented as a learning-free defense that operates during inference, which generates a counterattack perturbation $\vec{\delta}_{\text{ca}}$ that neutralizes potential adversarial perturbations in the input \cite{xing2025clip}. Formally, TTC maximizes the embedding distance between the adversarial example $\vec{x}_{\text{adv}}$ and the counterattack example $\vec{x}_{\text{ca}}=\vec{x}_{\text{adv}}+ \vec{\delta}_{\text{ca}}$ as
\begin{equation}\label{TTCOptimization}
    \vec{\delta}_{\text{ca}} = \arg\max_{\|\vec{\delta}_{\text{ca}}\|_p \leq \epsilon_{\text{ca}}} \| I_\theta(\vec{x}_{\text{adv}} + \vec{\delta}_{\text{ca}}) - I_\theta(\vec{x}_{\text{adv}}) \|,
\end{equation}
where $\epsilon_{\text{ca}}$ denotes the budget of counterattack perturbation. To approximate the maximization problem \eqref{TTCOptimization}, TTC adopts PGD to update counterattack perturbation $\vec{\delta}_{\text{ca}}$ as follows:
\begin{equation}\label{TTCPGD}
    \vec{\delta}^{t+1}_{\text{ca}}=\Pi\Big{[}\vec{\delta}^{t}_{\text{ca}}+\alpha\cdot\text{sign}\Big{(}\nabla_{\vec{x}_{\text{adv}}} \mathcal{L}\big{(}\vec{x}_{\text{adv}}, \vec{\delta}^{t}_{\text{ca}})\big{)}\Big{)}\Big{]},
\end{equation}
where $\mathcal{L}=\|I_\theta(\vec{x}_{\text{adv}}+\vec{\delta}_{\text{ca}})-I_\theta(\vec{x}_{\text{adv}})\big{)}\|$, $\Pi(\cdot)$ denotes the projection operation, and $\alpha$ signifies the step size.

\subsection{Directional Orthogonal Counterattack}
\subsubsection{Orthogonal Gradient Augmentation}
Crafting counterattacks using PGD presents a fundamental challenge due to the intrinsic differences between adversarial attacks and counterattacks. Specifically, while adversarial attacks maximize loss with respect to class labels as in equation \eqref{VLMAtk}, counterattacks operate without label supervision and aim to push the adversarial input away from its corrupted embedding as in equation \eqref{TTCOptimization}. On this basis, using PGD \eqref{TTCPGD}, which relies on gradients with respect to the adversarial input to generate counterattacks, restricts the optimization to a narrow region as defined in equation \eqref{TTCOptimization}, and fails to explore the adversarial space that truly requires neutralization, as described in equation \eqref{VLMAtk}. Furthermore, since ground-truth labels are unavailable at test time, addressing the mismatch in optimization objectives hinges critically on improving the counterattack strategy. Consequently, we propose enhancing the diversity of counterattacks to discover more generalizable solutions by exploring a broader region of adversarial space, which mitigates overfitting and better counteracts the underlying adversarial perturbation distribution.
\par
Therefore, we introduce randomized exploration along directions orthogonal to the primary gradient, coupled with the momentum-based update strategy. This design expands the counterattack search space, enabling it to escape narrow local optima and explore regions beyond the reach of standard PGD, thereby more effectively approximating and neutralizing a broader range of adversarial perturbations. As shown in Fig. \ref{Motivation} (c)-(d), DOC generates more dispersed and generalized counterattacks, guiding adversarial examples closer to the distribution of clean examples and enhancing robustness. Specifically, we first compute the normalized gradient:
\begin{equation}\label{GradDef}
    \vec{g} = \frac{\nabla_{\vec{x}_{\text{adv}}} \mathcal{L}\big{(}I_\theta(\vec{x}_{\text{adv}}+\vec{\delta}^{t}_{\text{ca}}),I_\theta(\vec{x}_{\text{adv}})\big{)}}{\| \nabla_{\vec{x}_{\text{adv}}} \mathcal{L}\big{(}I_\theta(\vec{x}_{\text{adv}}+\vec{\delta}^{t}_{\text{ca}}),I_\theta(\vec{x}_{\text{adv}})\big{)} \|}.
\end{equation}
Rather than updating solely along the gradient direction, we introduce an orthogonal component to expand the search region for counterattacks. Given the gradient \eqref{GradDef} and a vector $\vec{r} \sim \mathcal{N}(0, 1)$, we compute the orthogonal component as
\begin{equation}\label{OrthDef}
    \vec{r}^{\perp} = \frac{\vec{r} - \langle \vec{r}, \vec{g} \rangle \vec{g}}{\| \vec{r} - \langle \vec{r}, \vec{g} \rangle \vec{g} \|},
\end{equation}
where orthogonal projection ensures $\langle \vec{r}^{\perp}, \vec{g} \rangle = 0$. We then form the composite update direction $\vec{d}$ by combining the gradient direction and the orthogonal component as
\begin{equation}\label{UpdateDef}
    \vec{d} = \vec{g} + \lambda \cdot \vec{r}^{\perp},
\end{equation}
where $\lambda$ controls the strength of the orthogonal injection. To further alleviate the overfitting of counterattack perturbations and enhance their generalization, we adopt a momentum-based update scheme as follows:
\begin{equation}\label{momentum}
    \vec{m}_{t} = \mu \cdot \vec{m}_{t-1} + (1 - \mu) \cdot \vec{d},
\end{equation}
where $\mu \in [0, 1)$ is the momentum factor. Finally, the iterative role of our counterattack perturbation is presented as
\begin{equation}\label{CAtkOA}
    \vec{\delta}^{t+1}_{\text{ca}} = \Pi\big(\vec{\delta}^{t}_{\text{ca}} + \alpha \cdot \text{sign}(\vec{m}_{t}) \big).
\end{equation}
Compared to standard PGD, our method expands the counterattack search space and enhances perturbation diversity, enabling better generalization to a wider range of potential adversarial perturbations and thereby improving robustness.

\begin{algorithm}[t]
    \renewcommand{\algorithmicrequire}{\textbf{Input:}}
    \renewcommand{\algorithmicensure}{\textbf{Output:}}
    \caption{Implementation of DOC}
    \label{alg:doc}
    \begin{algorithmic}[1]
        \Require CLIP model $I_{\theta}$; Input example $\vec{x}$; Counterattack perturbation budget $\epsilon_{\text{ca}}$; Sample time $M$; Step size $\alpha$; Counterattack steps $T$; Hyperpatameters $\lambda$, $\tau$, and $\gamma$.
        \Ensure Counterattack perturbation $\vec{\delta}_{\text{ca}}$.
        \Statex {\centering {\texttt{/* Directional Sensitivity Score */}}\par}
        \For{$m = 1$ to $M$}
            \State $\vec{\eta}^m \gets \mathcal{U}(-\epsilon_{\text{ca}}, \epsilon_{\text{ca}})$.
            \State $\vec{x}^{m}_{\text{input}} = \vec{x}_{\text{input}} + \vec{\eta}^{m}$.
            \State $\tau_{\text{cos}} \gets \tau_{\text{cos}} + \cos\big{(} I_\theta(\vec{x}^{m}_{\text{input}}), I_\theta(\vec{x}_{\text{input}}) \big{)}$.
        \EndFor
        \State $\hat\tau(\vec{x}_{\text{input}})\gets1-\tau_{\text{cos}}/M$ as Eq. \eqref{SumCos}.
        \State $w \gets$ Eq. \eqref{DSDBinary}.
        
        \Statex {\centering {\texttt{/* Orthogonal Gradient Aug */}}\par}
        \State Initialize $\vec{m}_0 \gets 0$, $\vec{\delta}^0_{\text{ca}} \sim \mathcal{U}(-\epsilon_{\text{ca}}, \epsilon_{\text{ca}})$.
        \For{$t = 1$ to $T$}
            \State Normalized gradient $\vec{g}\gets$ Eq. \eqref{GradDef}.
            \State $\vec{r} \sim \mathcal{N}(0, 1)$.
            \State $\vec{r}^\perp \gets$ Eq. \eqref{OrthDef}.
            \State $\vec{d} \gets \vec{g} + \lambda \cdot \vec{r}^{\perp}$ as Eq. \eqref{UpdateDef}.
            \State $\vec{m}_t \gets \mu \cdot \vec{m}_{t-1} + (1 - \mu) \cdot \vec{d}$ as Eq. \eqref{momentum}.
            \State $\vec{\delta}^t_{\text{ca}} \gets \Pi\big(\vec{\delta}^{t}_{\text{ca}} + \alpha \cdot \text{sign}(\vec{m}_{t}) \big)$ as Eq. \eqref{CAtkOA}.
        \EndFor
        \State $\vec{\delta}_{\text{ca}}\gets w\cdot\vec{\delta_{\text{ca}}} + (1-w)\cdot \vec{\delta}^0_{\text{ca}}$.
    \end{algorithmic}
\end{algorithm}

\begin{table*}[t!]
    \centering
    \resizebox{\textwidth}{!}{
    \begin{tabular}[l]{@{}l| c| c| c c c c c c| c c c c | c |c}
    \toprule[2pt]
    \multirow{2}*{Dataset} &\multirow{2}*{Acc} &\multirow{2}*{CLIP} &\multicolumn{6}{c|}{Adversarial Fine-Tuning} &\multicolumn{4}{c|}{Test-Time Defence} &\multirow{2}*{$\Delta_o$} &\multirow{2}*{$\Delta_{\uparrow}$}\\
    \cmidrule(r){4-9} \cmidrule(r){10-13}
    & & &TeCoA$^1$ &TeCoA$^4$ &PMG$^1$ &PMG$^4$ &FARE$^1$ &FARE$^4$ &Anti &HD &TTC &DOC & &\\
    \toprule[1pt]
    \multirow{2}*{CIFAR10}     &Robust &0.00 &7.72 &11.83 &10.16 &15.79 &2.02 &5.47 &0.32 &1.82 &30.25 &\textbf{38.14} &38.14 &7.89\\
    &Clean &85.08 &64.64 &65.15 &70.68 &71.45 &74.46 &78.46 &\textbf{83.44} &78.23 &81.32 &81.25 &-3.83 &-2.19\\
    \toprule[1pt]
    \multirow{2}*{CIFAR100}    &Robust &0.00 &6.39 &9.39 &7.71 &11.12 &2.87 &4.59 &0.22 &0.96 &9.46 &\textbf{15.46} &15.46 &6.00\\
    &Clean &57.16 &35.94 &36.30 &40.32 &41.51 &46.67 &47.38 &53.96 &52.86 &\textbf{56.11} &55.96 &-1.20 &2.00\\
    \toprule[1pt]
    \multirow{2}*{STL10}       &Robust &0.04 &24.10 &31.91 &28.49 &35.77 &10.05 &17.72 &2.25 &3.80 &51.89 &\textbf{69.16} &69.12 &17.27\\
    &Clean &96.41 &87.40 &81.69 &88.56 &84.35 &91.76 &89.11 &95.47 &89.50 &\textbf{96.03} &95.83 &-0.58 &0.36\\
    \toprule[1pt]
    \multirow{2}*{ImageNet}    &Robust &0.00 &1.65 &3.07 &2.07 &3.71 &0.16 &0.83 &0.15 &0.04 &13.07 &\textbf{24.64} &24.64 &11.57\\
    &Clean &59.72 &34.89 &27.76 &36.12 &28.51 &48.79 &40.48 &54.29 &\textbf{54.54} &32.36 &41.91 &-17.81 &-12.63\\
    \toprule[1pt]
    \multirow{2}*{Caltech101}  &Robust &0.60 &15.70 &21.41 &19.50 &26.01 &5.14 &10.29 &3.14 &1.62 &35.90 &\textbf{52.05} &51.45 &16.15\\
    &Clean &85.69 &71.64 &64.41 &75.43 &69.06 &80.95 &76.58 &83.99 &82.33 &85.99 &\textbf{86.54} &0.85 &0.55\\
    \toprule[1pt]
    \multirow{2}*{Caltech256}  &Robust &0.13 &8.26 &12.14 &10.57 &13.88 &2.17 &5.39 &1.44 &0.55 &26.38 &\textbf{43.08} &42.95 &16.70\\
    &Clean &81.72 &61.11 &52.05 &62.20 &53.32 &73.28 &67.22 &\textbf{79.40} &79.12 &75.96 &79.24 &-2.48 &-0.16\\
    \toprule[1pt]
    \multirow{2}*{OxfordPets}  &Robust &0.00 &0.95 &3.96 &1.77 &5.19 &0.22 &0.32 &0.10 &0.00 &25.89 &\textbf{46.52} &46.52 &20.63\\
    &Clean &87.35 &62.06 &53.94 &65.85 &56.66 &79.37 &70.10 &80.53 &\textbf{80.91} &60.70 &74.05 &-13.30 &-6.86\\
    \toprule[1pt]
    \multirow{2}*{Flowers102}  &Robust &0.00 &1.84 &3.88 &2.55 &4.95 &0.03 &0.62 &0.05 &0.00 &13.77 &\textbf{27.99} &27.99 &14.22\\
    &Clean &65.43 &36.71 &27.78 &36.97 &28.88 &48.04 &41.01 &62.80 &58.22 &63.23 &\textbf{64.48} &-0.95 &1.25\\
    \toprule[1pt]
    \multirow{2}*{FGVCAircraft}&Robust &0.00 &0.03 &0.15 &0.03 &0.09 &0.00 &0.04 &0.00 &0.00 &7.77 &\textbf{11.19} &11.19 &3.42\\
    &Clean &20.07 &5.43 &3.51 &5.43 &3.24 &10.80 &7.77 &15.64 &16.36 &15.96 &\textbf{18.15} &-1.92 &1.79\\
    \toprule[1pt]
    \multirow{2}*{StanfordCars}&Robust &0.00 &0.15 &0.47 &0.15 &0.61 &0.01 &0.04 &0.00 &0.00 &12.66 &\textbf{24.57} &24.57 &11.91\\
    &Clean &52.07 &20.91 &15.18 &25.36 &16.79 &38.68 &32.09 &36.14 &44.28 &41.54 &\textbf{48.51} &-3.56 &4.23\\
    \toprule[1pt]
    \multirow{2}*{SUN397}      &Robust &0.00 &1.30 &2.31 &1.90 &3.37 &0.13 &0.65 &0.11 &0.00 &13.43 &\textbf{16.71} &16.71 &3.28\\
    &Clean &58.50 &36.69 &28.16 &37.98 &29.93 &52.42 &43.57 &\textbf{55.99} &53.17 &46.68 &47.15 &-11.35 &-8.84\\
    \toprule[1pt]
    \multirow{2}*{Country211}  &Robust &0.00 &0.05 &0.22 &0.12 &0.34 &0.00 &0.03 &0.00 &0.00 &2.72 &\textbf{4.98} &4.98 &2.26\\
    &Clean &15.22 &4.75 &3.66 &4.64 &3.34 &9.25 &6.58 &11.60 &11.72 &12.07 &\textbf{13.46} &-1.76 &1.39\\
    \toprule[1pt]
    \multirow{2}*{Food101}     &Robust &0.00 &0.56 &1.43 &1.03 &2.19 &0.06 &0.34 &0.07 &0.64 &18.52 &\textbf{34.74} &34.74 &16.22\\
    &Clean &83.86 &30.00 &21.90 &36.62 &27.97 &55.24 &41.98 &75.95 &80.30 &79.86 &\textbf{82.46} &-1.40 &2.16\\
    \toprule[1pt]
    \multirow{2}*{EuroSAT}     &Robust &0.00 &9.81 &10.82 &9.62 &10.52 &0.00 &7.58 &0.03 &0.49 &14.24 &\textbf{14.49} &14.49 &0.25\\
    &Clean &42.57 &16.36 &17.53 &18.14 &19.19 &21.10 &18.22 &36.81 &39.08 &\textbf{53.09} &52.92 &10.35 &-0.17\\
    \toprule[1pt]
    \multirow{2}*{DTD}         &Robust &0.11 &4.20 &5.19 &4.31 &5.30 &0.90 &2.89 &0.37 &0.16 &11.91 &\textbf{19.68} &19.57 &7.77\\
    &Clean &40.43 &25.16 &20.11 &21.76 &17.29 &31.97 &28.03 &\textbf{38.55} &34.89 &36.12 &36.44 &-3.99 &-2.11\\
    \toprule[1pt]
    \multirow{2}*{PCAM}        &Robust &0.00 &20.95 &44.13 &12.87 &36.38 &0.64 &3.74 &0.25 &12.04 &51.61 &\textbf{52.95} &52.95 &1.34\\
    &Clean &52.95 &49.96 &49.98 &12.87 &49.80 &52.53 &50.17 &52.61 &50.38 &53.11 &\textbf{53.84} &-0.89 &0.73\\
    \toprule[1pt]
    \toprule[1pt]
    \multirow{2}*{Average} &Robust  &0.06 &6.48 &10.15 &7.05 &10.95 &1.53 &3.78 &0.53 &1.38 &21.22 &\textbf{31.02} &30.96 &9.80\\
    &Clean &61.51 &40.23 &35.57 &39.93 &37.58 &50.96 &46.23 &57.32 &56.62 &55.63 &\textbf{58.26} &-3.25 &0.94\\
    \toprule[2pt]
\end{tabular}
}
\caption{Clean and robust accuracy under PGD-10 with $\epsilon_{\text{atk}} = 4/255$ on 16 datasets. Adversarial fine-tuning methods are trained on Tiny ImageNet, with superscripts indicating the attack budget used during fine-tuning. $\Delta_o$ indicates the improvement over the original CLIP, and $\Delta_{\uparrow}$ denotes the gain over the previous best. Bold indicates the best performance.}
\label{INResults}
\end{table*}

\subsubsection{Counterattack with Directional Sensitivity Score}
Counterattacks require identifying whether an input is a clean or an adversarial example to determine the need for countermeasures. Prior work addresses this by leveraging pseudo-stability, based on the observation that adversarial examples tend to exhibit larger embedding shifts under random perturbations \cite{wu2021attacking, xing2025clip}. This is measured by the $\ell_2$ distance between the input example and its noisy counterpart, but it raises two concerns. First, two embeddings may have similar directions but differ in scale, which can inflate the $\ell_2$ distance despite semantic similarity. Second, relying on a single noisy sample introduces randomness, making the decision process unstable.
\par
Correspondingly, we adopt cosine similarity to measure pseudo-stability, focusing on directional alignment and being invariant to scaling. Furthermore, we average the similarity over multiple random perturbations to mitigate stochastic effects and improve decision robustness. Specifically, for the input example $\vec{x}_{\text{input}}$ with unknown status as clean or adversarial, we generate $M$ noisy versions $\vec{x}^{m}_{\text{input}} = \vec{x}_{\text{input}} + \vec{\eta}^{m}$, where $\vec{\eta}^{m} \sim [\epsilon_\text{ca} \cdot \text{sign}(\mathcal{N}(0, 1))]$ as follows:
\begin{equation}\label{SumCos}
    \hat\tau(\vec{x}_{\text{input}}) = 1 - \frac{1}{M} \sum_{m=1}^{M} \cos \Big{(} I_\theta(\vec{x}^{m}_{\text{input}}), I_\theta(\vec{x}_{\text{input}}) \Big{)},
\end{equation}
where $\cos(\cdot,\cdot)$ denotes cosine similarity. A lower $\hat\tau(\vec{x})$ indicates that perturbed embeddings remain directionally aligned, suggesting the input is clean. Conversely, a higher score reflects directional inconsistency, indicating a potential adversarial example. To improve sample discriminability, we apply a soft gating function instead of a hard threshold, which avoids abrupt binary decisions and mitigates sensitivity to threshold hyperparameters as follows:
\begin{equation}\label{DSDBinary}
    w = \sigma \Big{(} \gamma \cdot \big{(}\tau - \hat\tau(\vec{x})\big{)}\Big{)} \in (0,1), 
\end{equation}
where $\tau$ denotes the predefined threshold, $\gamma$ controls the sharpness, and $\sigma(\cdot)$ is the sigmoid function. Therefore, the final counterattack perturbation $\vec{\delta_{\text{ca}}}$ is generated as $\vec{\delta_{\text{ca}}}=w\cdot\vec{\delta_{\text{ca}}} + (1-w)\cdot \vec{\delta}^0_{\text{ca}}$ with noise $\vec{\delta}^0_{\text{ca}} \sim \mathcal{U}(-\epsilon_{\text{ca}}, \epsilon_{\text{ca}})$.
\par
Compared to the $\ell_2$ norm, our directional sensitivity score based on cosine similarity provides more reliable indicators of adversarial perturbations, as it is less affected by irrelevant scaling in high-dimensional feature spaces. Meanwhile, rather than applying hard binarization, we employ an adaptive mechanism to modulate counterattack strength, enabling finer discrimination between inputs and more flexible responses. Additionally, averaging over multiple random perturbations mitigates the instability of single-sample estimates and improves the stability of counterattack decisions.

\section{Experiments and Analysis}
\subsection{Experiment Settings}
\textbf{Datasets for Evaluation}
We conduct systematic experiments and analyses across 16 datasets. For general object classification, we include CIFAR-10 / 100 \cite{krizhevsky2009learning}, STL-10 \cite{coates2011analysis}, ImageNet \cite{deng2009imagenet}, Caltech-101 \cite{fei2006one}, and Caltech-256 \cite{griffin2007caltech}. For fine-grained classification, we consider Oxford Pets \cite{parkhi2012cats}, Flowers-102 \cite{nilsback2008automated}, Food-101 \cite{bossard2014food}, and Stanford Cars \cite{krause20133d}. For scene recognition, we use SUN397 \cite{xiao2010sun} and Country211 \cite{CLIPRef}. In addition, we incorporate domain-specific datasets, including FGVC Aircraft \cite{maji2013fine}, EuroSAT \cite{helber2019eurosat}, DTD \cite{cimpoi2014describing}, and PatchCamelyon (PCAM) \cite{bejnordi2017diagnostic}.
\\
\begin{table*}[t!]
    \centering
    \resizebox{\textwidth}{!}{
    \begin{tabular}{c|*{16}{c}|c|c}
    \toprule[2pt]
    \rotatebox{90}{Method} & \rotatebox{70}{CIFAR10} & \rotatebox{70}{CIFAR100} & \rotatebox{70}{STL10} & \rotatebox{70}{ImageNet} &
    \rotatebox{70}{Caltech101} & \rotatebox{70}{Caltech256} & \rotatebox{70}{OxfordPets} &
    \rotatebox{70}{Flower102} & \rotatebox{70}{FGVCAircraft} & \rotatebox{70}{StanfordCars} &
    \rotatebox{70}{SUN397} & \rotatebox{70}{Country211} & \rotatebox{70}{Food101} &
    \rotatebox{70}{EuroSAT} & \rotatebox{70}{DTD} & \rotatebox{70}{PCAM} &
    \rotatebox{70}{Avg. Rob.} & \rotatebox{70}{Avg. Acc.} \\
    \toprule[1pt]
    CLIP    &0.00 &0.00 &0.03 &0.00 &0.07 &0.08 &0.00 &0.00 &0.00 &0.00 &0.00 &0.00 &0.00 &0.00 &0.11 &0.94 &0.07 &\textbf{61.51}\\
    HD      &1.68 &0.00 &1.71 &0.01 &0.23 &0.12 &0.00 &0.00 &0.00 &0.00 &0.00 &0.00 &0.02 &0.11 &0.07 &5.04 &0.56 &54.85\\
    TTC     &30.15 &8.64 &53.08 &11.98 &34.83 &25.15 &24.45 &12.85 &6.66 &11.38 &12.74 &2.21 &16.46 &14.66 &12.39 &52.07 &20.61 &55.63\\
    DOC     &\textbf{35.68} &\textbf{12.10} &\textbf{66.42} &\textbf{20.91} &\textbf{48.07} &\textbf{39.16} &\textbf{41.89} &\textbf{25.11} &\textbf{10.11} &\textbf{20.20} &\textbf{14.08} &\textbf{3.66} &\textbf{29.26} &\textbf{14.41} &\textbf{17.13} &\textbf{52.73} &\textbf{28.18} &58.34\\
    \toprule[1pt]
    $\Delta_{\text{CLIP}}$&35.68 &12.10 &66.39 &20.91 &48.00 &39.08 &41.89 &25.11 &10.11 &17.84 &13.87 &3.66 &29.26 &14.41 &17.02 &51.79 &27.69 &-3.17\\
    $\Delta_{\uparrow}$&5.53 &3.46 &13.34 &8.93 &13.24 &14.01 &17.44 &12.26 &3.45 &8.82 &1.34 &1.45 &12.80 &-0.25 &4.74 &0.66 &7.58 &2.71\\
\toprule[2pt]
\end{tabular}}
    \caption{Performance of DOC under CW attack with a perturbation budget of $\epsilon_{\text{atk}}=4/255$. $\Delta_{\text{CLIP}}$ denotes the improvement over the original CLIP, and $\Delta_{\uparrow}$ indicates the gain over the previous best performance. The best performance is shown in bold.}
\label{CWAA}
\end{table*}
\begin{figure*}[t]\centering
    \subfigure[Combined with TeCoA]{\includegraphics[scale=0.295]{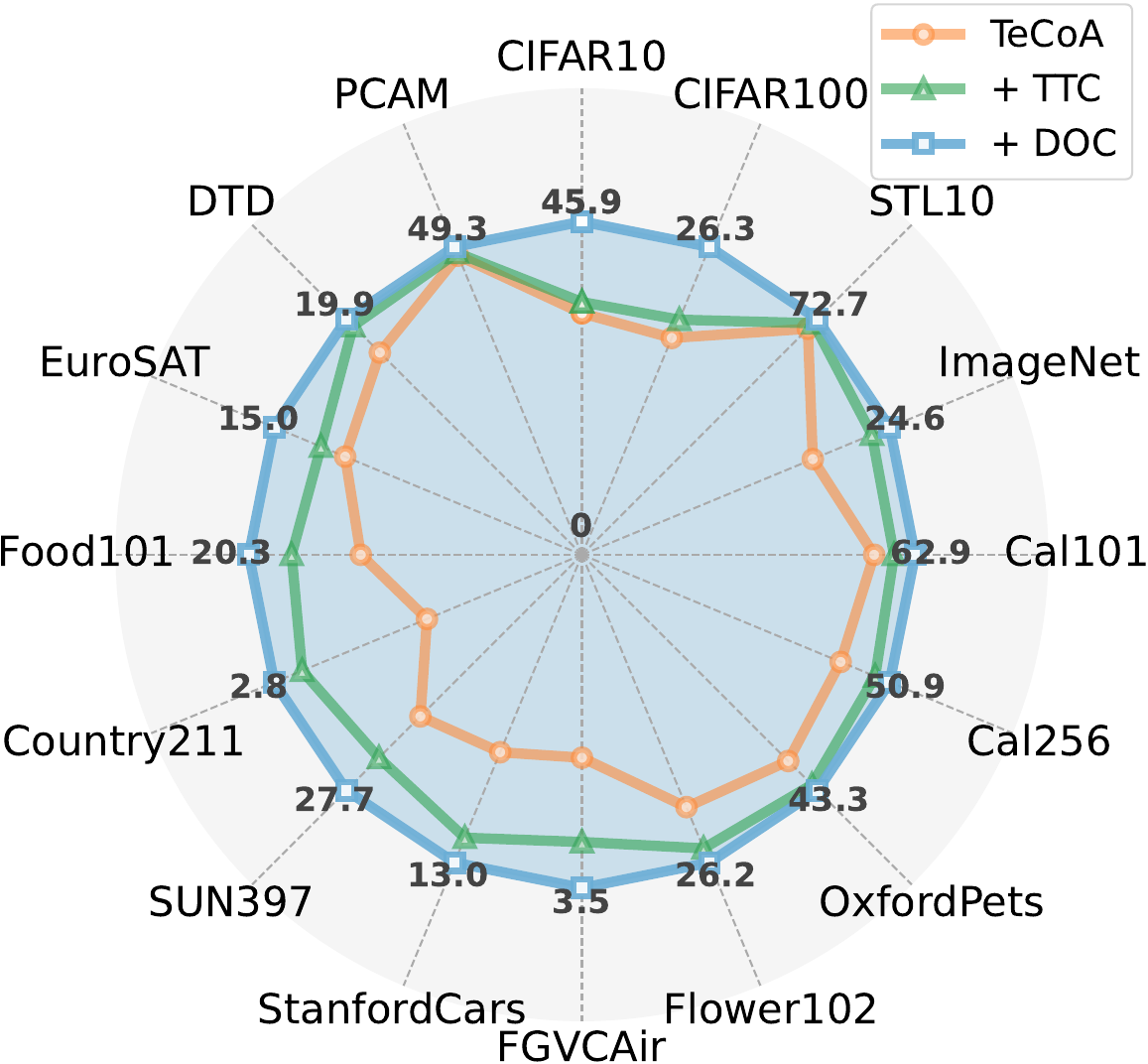}}
    \subfigure[Combined with PMG-AFT]{\includegraphics[scale=0.295]{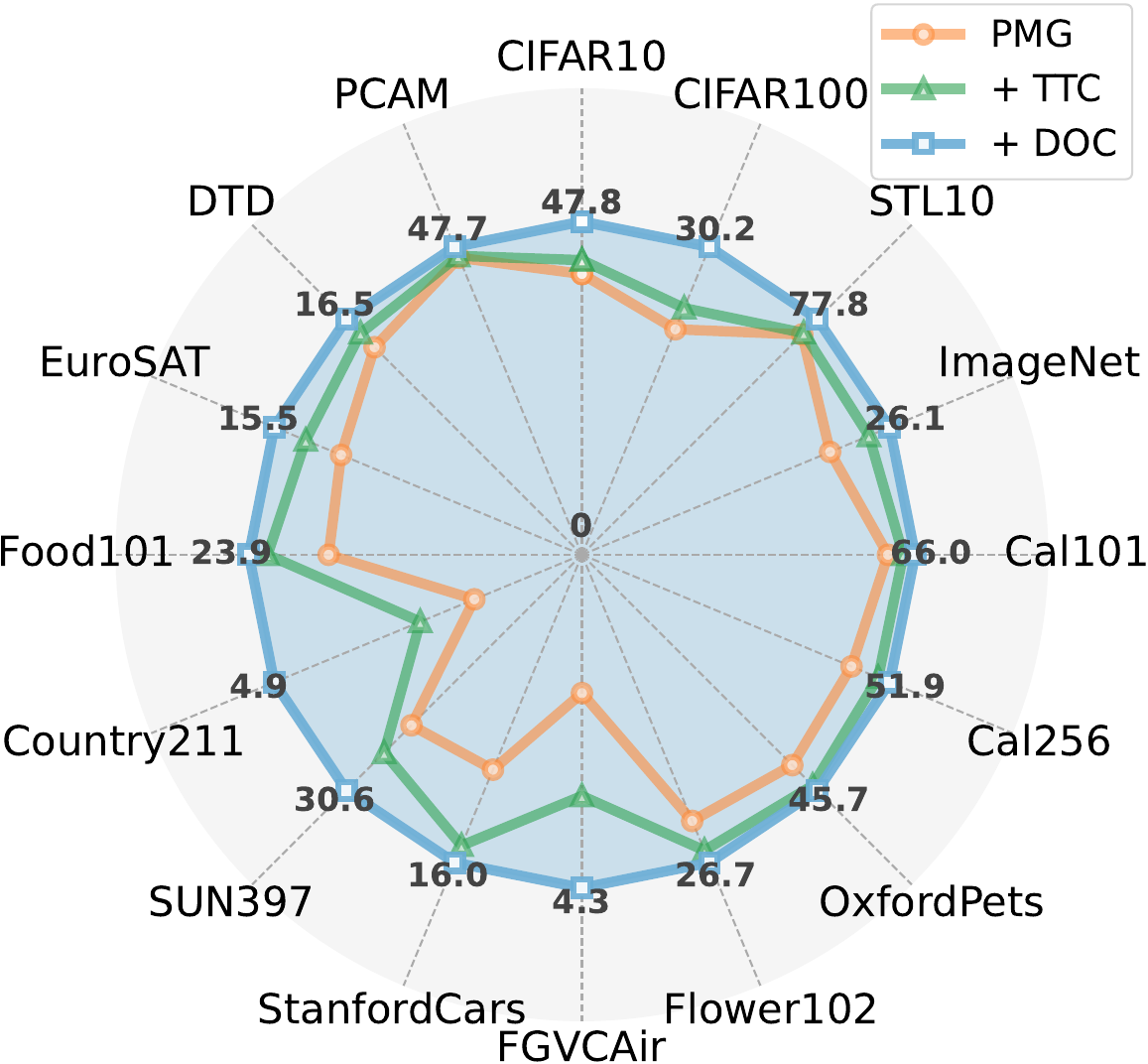}}
    \subfigure[Combined with FARE]{\includegraphics[scale=0.295]{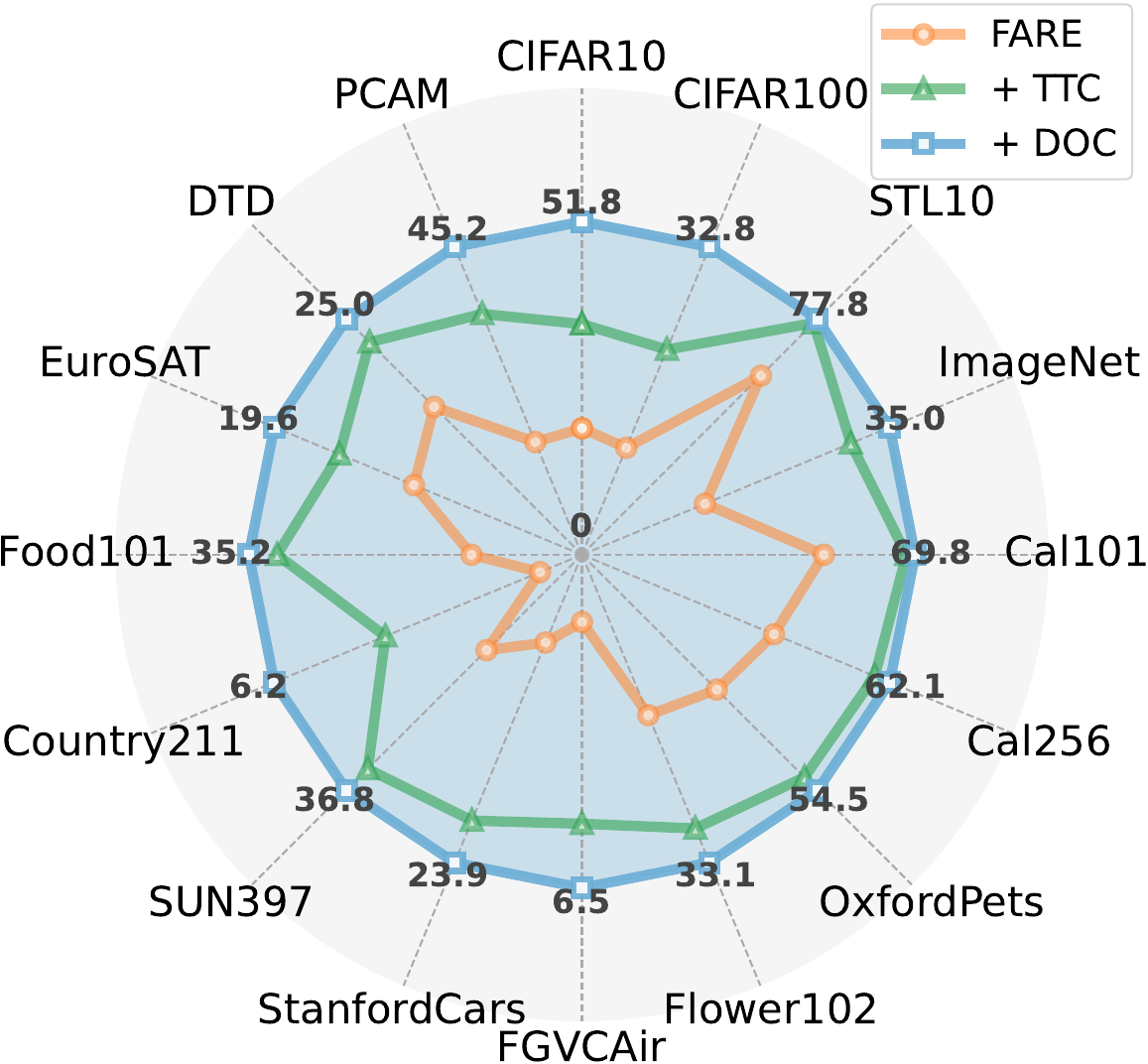}}
    \caption{Performance of DOC combined with adversarial fine-tuning, including TeCoA \cite{maounderstanding}, PMG-AFT \cite{wang2024pre}, and FARE \cite{schlarmann2024robust}. Robust accuracy is evaluated on 16 datasets using PGD-10 with $\epsilon_{\text{atk}} = 1/255$.}
    \label{CombineAFT}
\end{figure*}
\textbf{Baselines for Comparison}
As research on improving the zero-shot adversarial robustness of VLPs via test-time defense is still in its early stages and available methods are limited, we primarily compare our DOC with the state-of-the-art approach, TTC \cite{xing2025clip}. We further include representative test-time defenses, covering Anti-Adversary (Anti) \cite{alfarra2022combating} and Hedge Defense (HD) \cite{wu2021attacking}. Although our method targets test-time defense, we also compare it with three adversarial fine-tuning approaches, including TeCoA \cite{maounderstanding}, PMG-AFT (PMG) \cite{wang2024pre}, and FARE \cite{schlarmann2024robust}, which fine-tune CLIP on Tiny ImageNet.
\\
\textbf{Implementation Details}
The counterattack budget is set to $\epsilon_{\text{ca}} = 4/255$, following prior work \cite{xing2025clip}. We evaluate adversarial robustness under PGD \cite{madry2018towards}, CW \cite{carlini2017towards}, and AutoAttack (AA) \cite{croce2020reliable} with $\epsilon_{\text{atk}} = 4/255$ bounded by $\ell_\infty$ norm. The counterattack is performed with a batch size of 256 and 4 steps using a default step size of $\alpha_{\text{ttc}} = 3/255$. All experiments are conducted on a single NVIDIA 4090 GPU. Additional results under alternative settings are provided in the \textbf{Supplementary Materials}. 

\subsection{Main Results}
\subsubsection{Adversarial Robustness under PGD}
We evaluate our method and baselines under PGD-10 across 16 datasets, and the results are shown in Table \ref{INResults}. While adversarial fine-tuning methods improve robustness, they significantly degrade clean accuracy, and this degradation becomes more severe as the fine-tuning perturbation budget increases. Moreover, adversarial fine-tuning requires access to source data and incurs additional computational overhead. In contrast, our DOC achieves significant improvements in adversarial robustness while maintaining competitive clean accuracy. Specifically, DOC outperforms the state-of-the-art TTC, improving the average robust accuracy by $9.80\%$, and retains a higher clean accuracy. Furthermore, compared to the original CLIP model, DOC improves robust accuracy by over $30\%$ with minimal impact on clean performance, demonstrating its competitiveness as a test-time defense. 
\begin{figure}[t]\centering
    \includegraphics[scale=0.65]{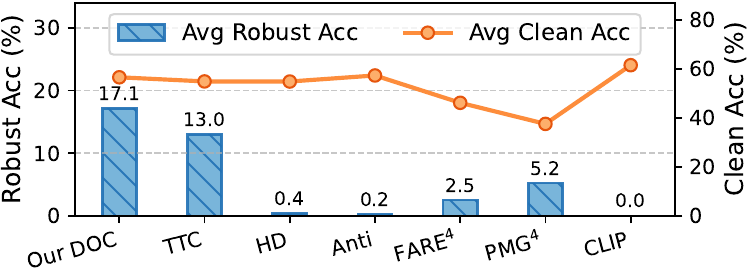}
    \caption{Performance of DOC and other baselines under AutoAttack with a perturbation budget of $\epsilon_{\text{atk}} = 4/255$. Clean and robust accuracy is averaged across 16 datasets.}
    \label{AAAcc}
\end{figure}

\begin{figure*}[t]\centering
    \subfigure[CIFAR10]{\includegraphics[scale=0.34]{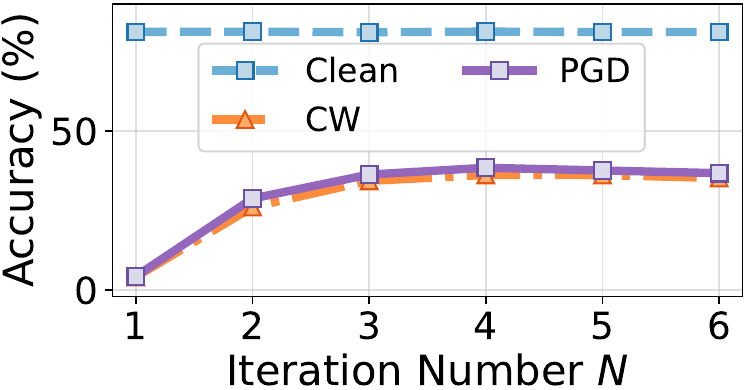}}
    \subfigure[CIFAR100]{\includegraphics[scale=0.34]{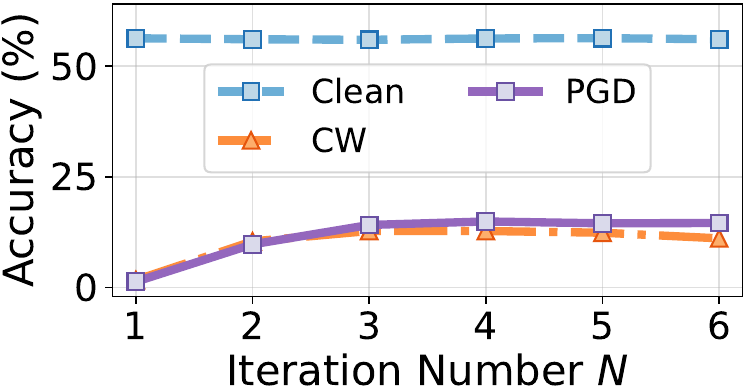}}
    \subfigure[STL10]{\includegraphics[scale=0.34]{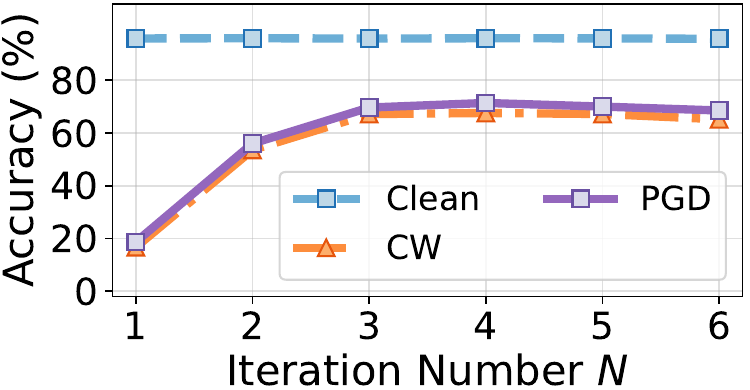}}
    \subfigure[ImageNet]{\includegraphics[scale=0.34]{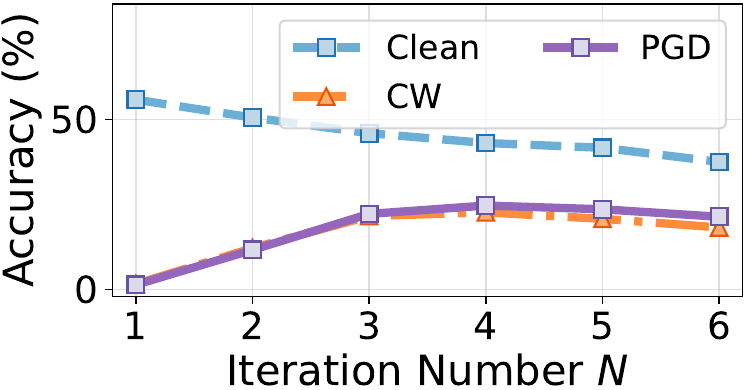}}
    \subfigure[Caltech101]{\includegraphics[scale=0.34]{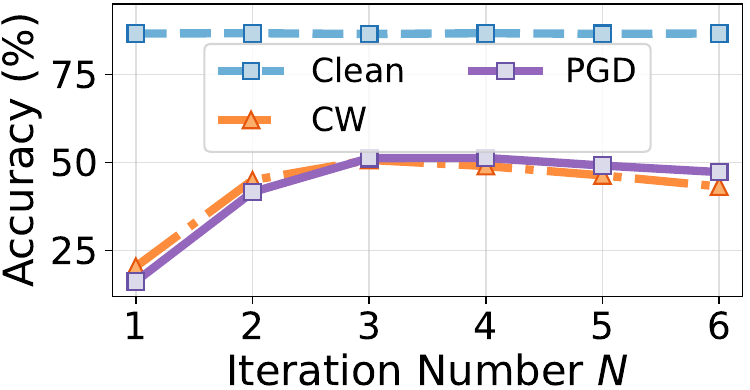}}
    \subfigure[Caltech256]{\includegraphics[scale=0.34]{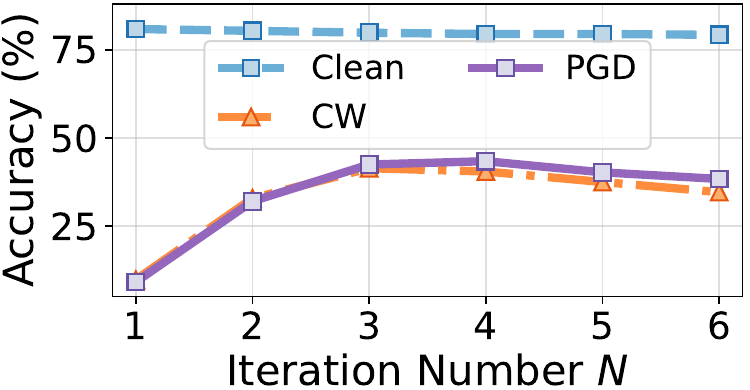}}
    \subfigure[OxfordPets]{\includegraphics[scale=0.34]{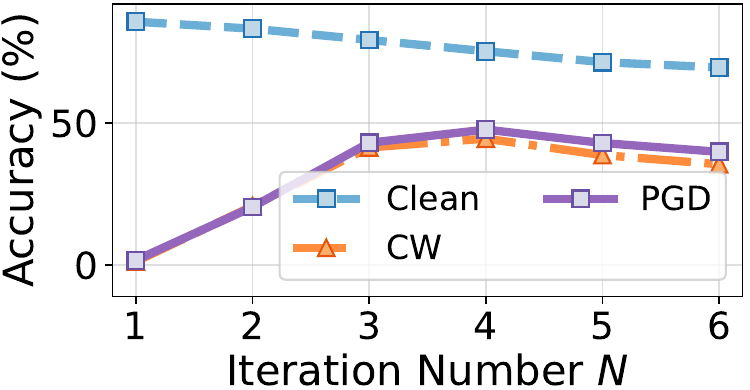}}
    \subfigure[Flowers102]{\includegraphics[scale=0.34]{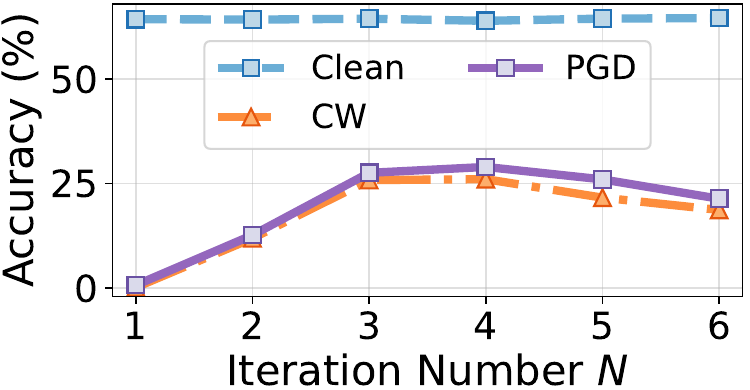}}
    \subfigure[FGVCAircraft]{\includegraphics[scale=0.34]{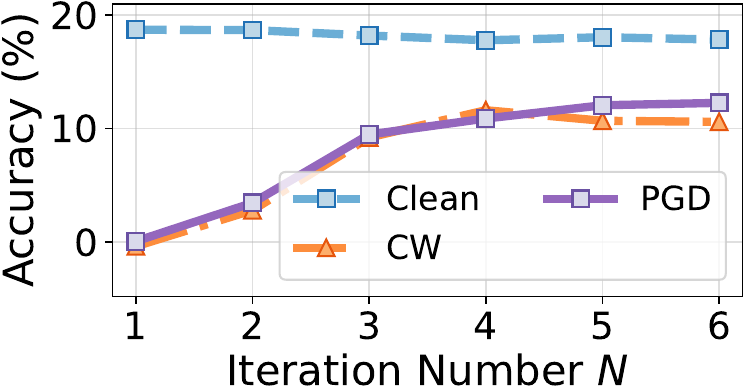}}
    \subfigure[StanfordCars]{\includegraphics[scale=0.34]{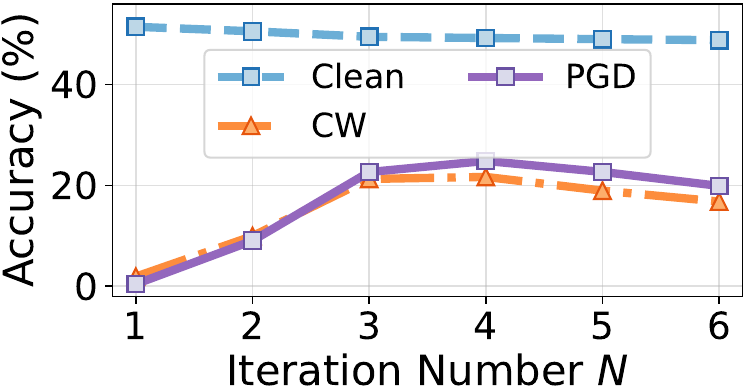}}
    \subfigure[Country211]{\includegraphics[scale=0.34]{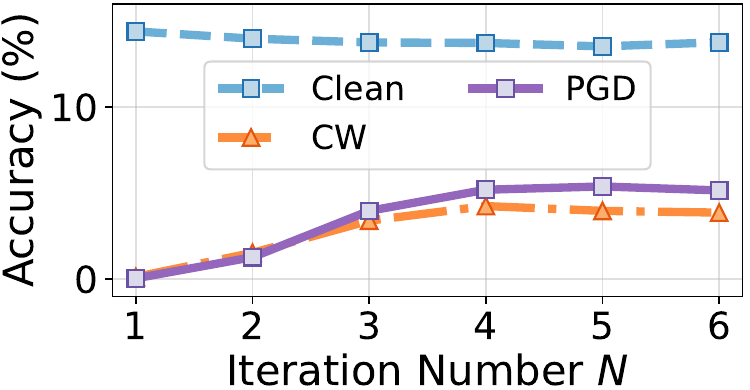}}
    \subfigure[Food101]{\includegraphics[scale=0.34]{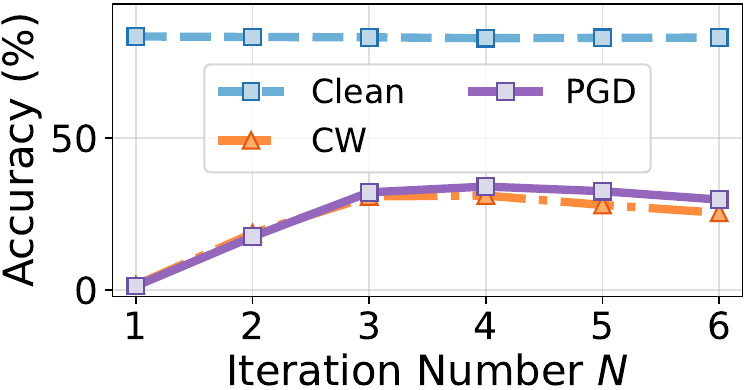}}
    \caption{Performance with numbers of counterattack steps $N$ on different datasets. Robust accuracy is evaluated by PGD-10 and CW with the perturbation budget $\epsilon_{\text{atk}} = 4/255$. Results on remaining datasets are presented in \textbf{Supplementary Materials}.}
    \label{AtkStep}
\end{figure*}

\subsubsection{Adversarial Robustness under CW and AutoAttack}
We further evaluate the robustness of our DOC against stronger attacks, including CW and AutoAttack. The corresponding results are reported in Table \ref{CWAA} (CW) and Fig. \ref{AAAcc} (AutoAttack). Specifically, our method consistently outperforms prior approaches, achieving average improvements of over $7.58\%$ under CW and $4.1\%$ under AutoAttack across 16 datasets. Compared to TTC, which also leverages CLIP’s pretrained features for counterattack generation, DOC introduces enhancements such as directional sensitivity discrimination and orthogonal-guided optimization, leading to consistent and better defense performance, with gains observed on nearly all datasets. Importantly, these improvements are achieved without additional training costs, making DOC practical for real-world deployment.

\subsection{Combining DOC with Adversarial Fine-Tuning}
Although our DOC is designed as a test-time defense, it can be integrated as a plug-in module to further enhance adversarially fine-tuned models. We follow the settings in \cite{xing2025clip} and report the results in Fig. \ref{CombineAFT}. When applied to adversarially finetuned models, covering TeCoA, PMG-AFT, and FARE, DOC consistently improves adversarial robustness, which brings an improvement of $4\%-5\%$ over the baselines. Notably, when combined with FARE, DOC achieves an average robust accuracy increase of over $18\%$ compared to the original CLIP. Interestingly, we observe that the magnitude of robustness gains varies across fine-tuning methods. This likely stems from the fact that adversarial fine-tuning can reduce the model’s embedding space sensitivity to input perturbations, which, while improving robustness, may also compromise the representational adopted for effective counterattack generation. Despite this, DOC remains effective in most cases, underscoring its adaptability and ability to leverage both pre-trained and fine-tuned encoder representations. Overall, DOC can serve as a lightweight enhancement to adversarial fine-tuning, without introducing additional training costs.

\subsection{Ablation Study}
\begin{table}[t]
    \centering
    \setlength{\tabcolsep}{0.5mm}{
    \begin{tabular}[l]{@{}c c| c| c| c| c}
    \toprule[2pt]
    \multirow{1}*{DSS} &\multirow{1}*{OGA} &Clean &PGD &CW &AA\\
    \toprule[1pt]
    \xmark &\xmark &55.66$_{\pm0.08}$ &21.43$_{\pm0.07}$ &20.70$_{\pm0.11}$ &21.97$_{\pm0.16}$\\
    \cmark &\xmark &58.23$_{\pm0.05}$ &23.37$_{\pm0.06}$ &22.27$_{\pm0.07}$ &22.66$_{\pm0.11}$\\
    \xmark &\cmark &55.38$_{\pm0.12}$ &31.83$_{\pm0.10}$ &29.02$_{\pm0.12}$ &26.07$_{\pm0.19}$\\
    \cmark &\cmark &58.27$_{\pm0.09}$ &31.04$_{\pm0.08}$ &28.15$_{\pm0.13}$ &25.89$_{\pm0.18}$\\
    \toprule[2pt]
    \end{tabular}
}
\caption{Ablation study results of our DOC. Clean and robust accuracy is reported as the average across 16 datasets. DSS and OGA denote the directional sensitivity score and the orthogonal gradient augmentation, respectively.}
\label{Ablation}
\end{table}
We conduct ablation experiments to evaluate the contribution of each component in DOC. Table \ref{Ablation} reports the average clean and robust accuracy across 16 datasets under $\epsilon_{\text{atk}} = 4/255$ with five random seeds (1-5). We adopt TTC as the baseline. Enabling DSS alone improves clean accuracy over the baseline by better distinguishing between clean and adversarial examples, which suppresses unnecessary perturbations on clean inputs, reducing the risk of amplifying benign variations into adversarial directions. Using OGA alone yields larger gains in robust accuracy, supporting our design motivation that diversity counterattack directions help neutralize adversarial perturbations more effectively without supervised information. Combining DSS and OGA achieves the best balance by improving both robustness and clean accuracy, which confirms DOC provides a reliable discrimination mechanism to prevent over-correction on clean examples and better neutralize adversarial perturbations. 

\subsection{Hyperparameter Selection and Discussion}
Due to page limitations, we analyze the key hyperparameter, counterattack steps $N$, while results for other parameters are included in the \textbf{Supplementary Materials}. Specifically, we use the default settings and an adversarial perturbation budget of $\epsilon_{\text{atk}} = 4/255$. As shown in Fig. \ref{AtkStep}, increasing $N$ consistently improves robustness up to $N = 3$ and saturates around $N = 3$ or $N = 4$. This trend suggests that even a small number of counterattack steps can yield substantial adversarial robustness gains, and that selecting an appropriate $N$ enables sufficient exploration of the adversarial perturbation space to effectively suppress adversarial effects. Importantly, clean accuracy remains stable, confirming that our DOC improves robustness not by sacrificing clean performance. The consistent robustness gains across both low-resolution and fine-grained datasets certify the competitiveness of our DOC in improving adversarial robustness.

\section{Conclusion and Future}
This work revisits the optimization strategy for counterattacks in test-time defense and identifies that vanilla PGD-based updates lack perturbation diversity, limiting their effect in neutralizing diverse adversarial patterns. Accordingly, we present Directional Orthogonal Counterattack (DOC), which enhances diversity by expanding the perturbation space through orthogonal exploration and momentum-based optimization, thereby better counteracting potential adversarial perturbation. In addition, DOC incorporates a directional sensitivity score computed via averaged cosine similarity, offering a stable and more discriminative criterion to adaptively guide counterattack strength. 
\par
Although developed on CLIP, our method does not rely on specific network architectures, label supervision, or training data. Instead, our DOC exploits the model’s intrinsic representational capacity, enabling straightforward transfer to other multimodal systems, including large-scale vision-language models. More importantly, we show that enhancing counterattack diversity substantially improves adversarial robustness, offering a promising direction for lightweight and scalable multimodal defenses.

\section{Acknowledgments}
This work was supported in part by the grant of the National Natural Science Foundation of China under Grant 62172090, in part by the Start-up Research Fund of Southeast University under Grant RF1028623097, in part by the Start-up Grant (No. 9610680) of the City University of Hong Kong, and in part by the Young Scientist Fund (No. 62406265) of NSFC. We thank the Big Data Computing Center of Southeast University for providing the facility support on the numerical calculations.

\bibliography{aaai2026}
\clearpage


\section*{Supplementary material}

# Supplementary Material of "Diversifying Counterattacks: Orthogonal Exploration for Robust CLIP Inference"

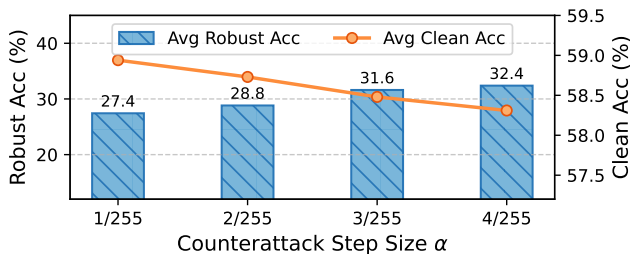

Figure 1: Effect of counterattack step size $\alpha$ on clean and robust accuracy across 16 datasets. Robustness is evaluated by PGD-10 with perturbation budget $\epsilon_{\text{atk}} = 4/255$.

## More Experimental Results and Analysis

### Experimental Environment Setup Details

All experiments are conducted on machines equipped with NVIDIA 4090 GPUs, using PyTorch 3.9.13 and CUDA 12.0. We adopt the publicly available CLIP model ViT-B/32 provided in Hugging Face as the backbone, and freeze model parameters throughout the evaluation to ensure a consistent inference-only setting (Radford et al. 2021). We use Projected Gradient Descent (PGD) (Madry et al. 2018), Carlini-Wagner (CW) (Carlini and Wagner 2017), and AutoAttack (Croce and Hein 2020) to evaluate adversarial robustness. The perturbation budget is set to $\epsilon_{\text{atk}} = 4/255$, and PGD is performed with 10 steps and a step size of $\alpha = 1/255$. All evaluations are conducted under a fixed random seed (1) to ensure reproducibility.

### Robustness under PGD with $\epsilon_{\text{atk}} = 1/255$

We evaluate the robustness of all methods under the adversarial perturbation budget of $\epsilon_{\text{atk}} = 1/255$. Following standard protocol in CLIP robustness studies, we perform PGD-10 attacks on 16 datasets and report the results in Table 1. Finetuning-based defenses yield moderate improvements in robustness but often incur noticeable drops in clean accuracy, highlighting their tradeoff between accuracy and robustness. Among test-time defenses, Anti-adversary and HD generate additive perturbations guided by handcrafted objectives, but their effectiveness is limited, leading to only marginal gains in robust accuracy. TTC leverages the pretrained CLIP encoder to optimize counterattacks, resulting

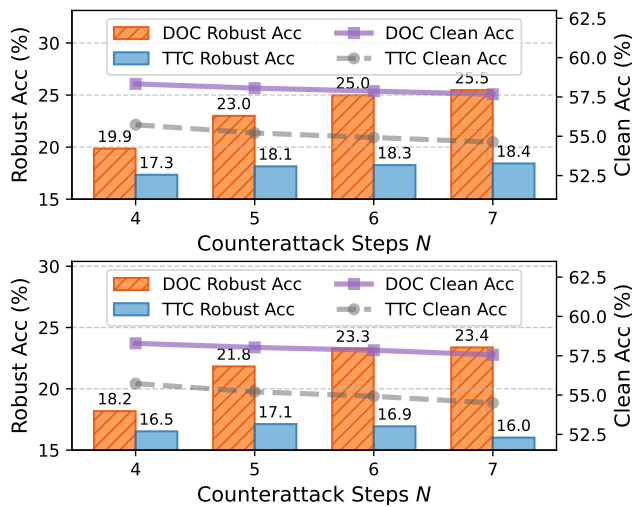

Figure 2: Robust and clean accuracy under a higher adversarial attack perturbation budget with $\epsilon_{\text{atk}} = 8/255$, including PGD-10 (top) and CW (bottom) attacks, evaluated with varying numbers of counterattack steps.

in stronger robustness than prior methods. However, its reliance on local gradient directions restricts the diversity of counterattack perturbations, limiting its ability to neutralize a broad spectrum of adversarial patterns. In contrast, our proposed DOC improves perturbation diversity by introducing orthogonal exploration and momentum-based updates, allowing it to discover more generalizable counterattacks and better defend against diverse adversarial inputs. DOC achieves the highest robust accuracy across most downstream datasets while maintaining competitive clean accuracy. Compared to the original CLIP, DOC improves average robust accuracy by $+43.31\%$, with only a slight clean accuracy reduction, demonstrating its competitiveness as a test-time defense under low-budget attacks.

### Robustness under PGD and CW with $\epsilon_{\text{atk}} = 8/255$

To further evaluate the competitiveness of our DOC under stronger adversarial threats, we conduct comparisons against PGD-10 and CW attacks with increased perturbation budgets $\epsilon_{\text{atk}} = 8/255$. The counterattack budget is consisted

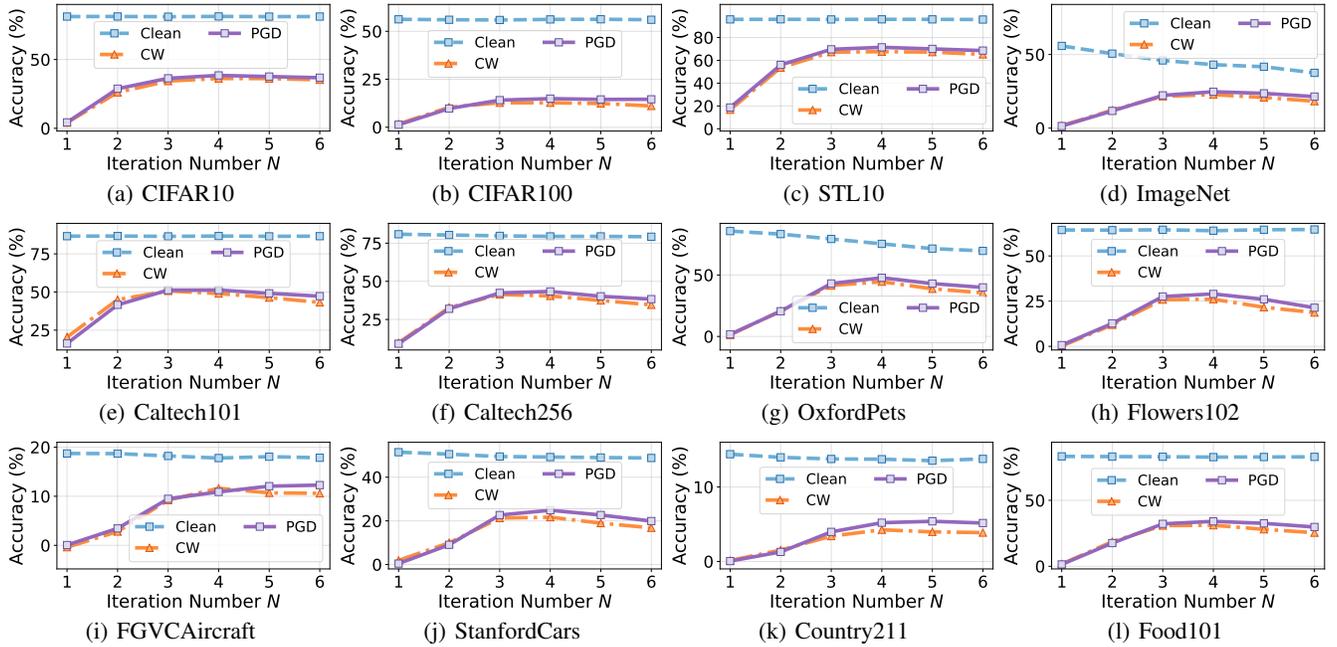

Figure 3: Performance of our presented DOC with different numbers of counterattack steps $N$. Robustness is evaluated by PGD-10 and CW attack with the adversarial perturbation budget $\epsilon_{\text{atk}} = 4/255$.

as $\epsilon_{\text{ca}} = 4/255$. As shown in Fig. 2, our method outperforms TTC in terms of robust accuracy while maintaining better clean accuracy. Specifically, DOC exhibits clear robustness gains over TTC as the number of counterattack steps increases from 4 to 7, which verifies that the orthogonal and momentum-driven design of DOC allows more thorough exploration of the adversarial perturbation space, resulting in stronger defense against attacks with high budgets. Similarly, under CW attacks, DOC continues to maintain superior robustness while preserving clean accuracy, which confirms that DOC exhibits more stable and generalizable behavior under different threat models. These results validate our hypothesis that enhancing counterattack diversity and exploration capability is essential to resisting adaptive and strong adversarial attacks. The consistent robustness improvement across both PGD and CW attacks demonstrates the generalizability and scalability of the proposed DOC.

### Effects of Counterattack Step-Size

We investigate how the counterattack step size affects the clean and robust performance of DOC. Specifically, we vary the $\ell_\infty$ norm of the per-step perturbation magnitude from $1/255$ to $4/255$ and report average clean and robust accuracy across 16 datasets. We adopt the PGD-10 with perturbation budget $\epsilon_{\text{atk}} = 4/255$ to evaluate robustness. As shown in Fig. 1, as the step size increases, DOC exhibits a clear upward trend in robust accuracy, improving from $27.43\%$ at $1/255$ to $32.39\%$ at $4/255$. This indicates that stronger per-step perturbations enable DOC to traverse a broader region of the embedding space, enhancing its ability to neutralize adversarial inputs. However, we observe a slight decrease in clean accuracy as the step size increases, dropping from $58.94\%$ to $58.31\%$. This trade-off suggests that while larger step sizes improve robustness, they potentially affect clean predictions. Nevertheless, the clean accuracy remains relatively stable, with less than a $0.7\%$ difference across all settings. Overall, a moderate-to-large step size such as $3/255$ or $4/255$ provides a favorable balance, delivering substantial robustness gains with minimal clean accuracy degradation.

### Effects of Counterattack Step Number

To complement the analysis in the main manuscript, we present detailed results on the impact of counterattack step number N across 12 diverse datasets, as shown in Fig. 3, which verifies the generalizability of our observations in the main manuscript and further supports the choice of number of steps for balancing robustness and efficiency. Consistent with Figure 6 in the main manuscript, we observe that adversarial robustness improves substantially as N increases from 1 to 3, with the performance plateauing around N=3 or N=4. This behavior validates our hypothesis that a moderate number of counterattack steps is sufficient to expand the perturbation space and increase the diversity of counterattacks. Moreover, as shown in all subplots, clean accuracy remains stable regardless of the choice of counterattack step, indicating DOC does not trade off clean performance for robustness. This confirms that the counterattack updates are well-regularized, especially with the help of momentum and orthogonal exploration, which allow the method to generalize without overfitting to specific perturbation modes.

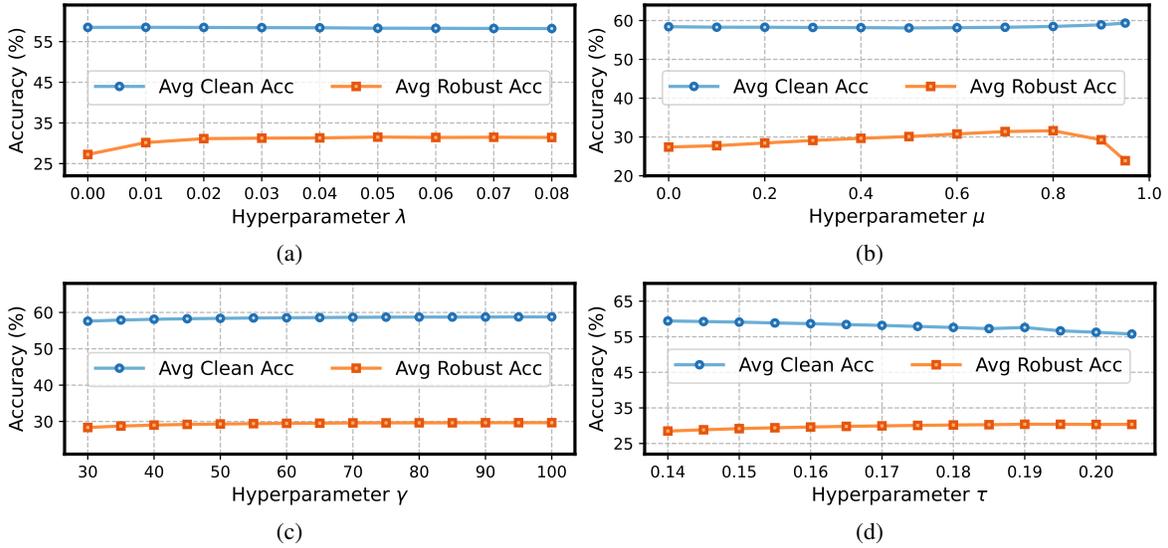

Figure 4: Effect of DOC hyperparameters. Clean and robust accuracy (%) is reported as an average over 16 datasets. Robustness is evaluated using PGD-10 under the adversarial perturbation budget of $\epsilon = 4/255$

## Hyperparameter Selection and Discussion

We conduct the hyperparameter analysis to determine how each hyperparameter influences the clean accuracy and robustness of the presented DOC. Specifically, our DOC includes four key hyperparameters: the orthogonal component factor ($\lambda$), the momentum factor ($\mu$), the discrimination threshold ($\tau$), and the sharpness factor ($\gamma$). To isolate the effect of each hyperparameter, we adopt a control-variable strategy, varying one parameter at a time while keeping the others fixed. The objective is to identify configurations that improve adversarial robustness without compromising clean accuracy. The analysis proceeds sequentially: we first tune the orthogonal component factor $\lambda$, followed by the momentum factor $\mu$, the discrimination threshold $\tau$, and finally the sharpness factor $\gamma$. We adopt PGD-10 with adversarial perturbation budget $\epsilon_{\text{atk}} = 4/255$ to generate adversarial examples for evaluating the adversarial robustness of CLIP, and report the average accuracy on 16 datasets.

**Effect of Orthogonal Component Factor $\lambda$** The orthogonal component factor $\lambda$ controls the strength of the randomized direction added orthogonally to the primary gradient during counterattack optimization. This component is designed to increase the diversity of the counterattack trajectory, helping it escape narrow local optima and enhance diversity. As shown in Fig. 4(a), increasing $\lambda$ from 0.00 to 0.05 steadily improves robustness, with robust accuracy rising from 27.25% to a peak of 31.52% at $\lambda = 0.05$. This trend validates our motivation that moderate orthogonal exploration improves the generalization of counterattacks by enhancing their diversity. Meanwhile, clean accuracy remains stable throughout the range, fluctuating slightly between 58.2% and 58.5%. When $\lambda$ exceeds 0.05, robustness begins to decline, likely due to excessive deviation from the primary gradient direction, which introduces instability into the optimization process. Overall, setting $\lambda$ within the range of 0.04 to 0.06 achieves a favorable balance between robustness and performance.

**Effect of Momentum Factor $\mu$** The momentum factor $\mu$ accumulates historical gradient information during counterattack optimization, helping the counterattack escape local optima and improving its generalization. As shown in Fig. 4(b), robustness improves as $\mu$ increases from 0.0 to 0.8, rising from 27.41% to a peak of 31.61%. Notably, clean accuracy also remains stable throughout this range, suggesting that the momentum-enhanced updates do not compromise standard performance. When $\mu$ exceeds 0.8, robustness begins to degrade and drops sharply at $\mu = 0.95$, which indicates that overly large momentum may cause overaccumulation of gradients, pushing counterattacks toward unstable or suboptimal directions. Thus, moderate values of $\mu$ (e.g., 0.7 to 0.8) offer a favorable trade-off, enabling DOC to maintain stable optimization while increasing diversity and generalization of counterattacks.

**Effect of Sharpness Factor $\gamma$** The sharpness factor $\gamma$ controls how sharply the counterattack strength responds to the estimated directional sensitivity score. A larger $\gamma$ increases the response sensitivity, allowing the counterattack strength to adapt more distinctly across samples with different adversarial characteristics. As shown in Fig. 4(c), both clean and robust accuracy steadily improve as $\gamma$ increases from 30 to 100. Specifically, robust accuracy rises from 28.34% at $\gamma = 30$ to 29.66% at $\gamma = 100$, while clean accuracy improves from 57.61% to 58.78%. This trend suggests that sharper modulation of counterattack strength improves sample-wise adaptation and overall robustness. Since the performance gains gradually plateau beyond $\gamma = 80$, further increasing $\gamma$ yields marginal benefits. In summary, a sharpness factor in the range of 70 to 100 offers strong robustness

| Dataset | Acc | CLIP | Adversarial Fine-Tuning | | | Test-time Defence | | | | | $\Delta_{\text{CLIP}}$ | $\Delta_\uparrow$ |
|---|---|---|---|---|---|---|---|---|---|---|---|---|
| | | | TeCoA[1] | PMG[1] | FARE[1] | RN | Anti-adv | HD | TTC | DOC (Ours) | | |
| CIFAR10 | Robust | 0.66 | 33.67 | 40.71 | 19.65 | 2.01 | 12.39 | 17.22 | 29.20 | **47.78** | 47.12 | 7.07 |
| | Clean | 85.09 | 64.64 | 70.68 | 74.46 | 81.18 | **83.52** | 78.23 | 81.32 | 81.99 | -3.10 | -1.53 |
| CIFAR100 | Robust | 0.21 | 18.93 | 22.55 | 11.41 | 0.67 | 5.73 | 3.86 | 15.36 | **24.35** | 24.14 | 1.80 |
| | Clean | 57.16 | 35.94 | 40.32 | 46.67 | 56.34 | 53.95 | 52.86 | 56.11 | **56.62** | -0.54 | 0.51 |
| STL10 | Robust | 11.47 | 70.09 | 73.11 | 59.10 | 16.23 | 37.42 | 39.02 | 76.58 | **86.33** | 74.86 | 9.22 |
| | Clean | 96.41 | 87.40 | 88.56 | 91.76 | 95.85 | 95.45 | 89.50 | 96.03 | **96.04** | -0.37 | 0.19 |
| ImageNet | Robust | 1.20 | 18.89 | 21.43 | 14.00 | 1.77 | 8.67 | 6.63 | 39.22 | **43.72** | 42.52 | 4.50 |
| | Clean | 59.73 | 34.89 | 36.12 | 48.79 | **59.34** | 54.27 | 54.54 | 46.12 | 46.46 | -13.26 | -12.88 |
| Caltech101 | Robust | 14.64 | 55.55 | 61.06 | 50.67 | 18.90 | 34.81 | 31.53 | 63.25 | **71.30** | 56.66 | 8.05 |
| | Clean | 85.70 | 71.64 | 75.43 | 80.95 | 86.61 | 84.02 | 82.33 | 86.37 | **88.09** | 2.39 | 1.48 |
| Caltech256 | Robust | 8.41 | 43.20 | 45.88 | 38.74 | 11.33 | 25.36 | 23.48 | 57.58 | **65.93** | 57.52 | 8.35 |
| | Clean | 81.73 | 61.11 | 62.20 | 73.28 | **81.25** | 79.38 | 79.12 | 78.14 | 79.45 | -2.28 | -1.80 |
| OxfordPets | Robust | 1.17 | 38.29 | 41.18 | 31.18 | 1.86 | 20.42 | 12.04 | 61.21 | **67.18** | 66.01 | 5.97 |
| | Clean | 87.33 | 62.06 | 65.85 | 79.37 | **87.41** | 80.62 | 80.91 | 78.17 | 81.36 | -5.97 | -6.05 |
| Flowers102 | Robust | 1.01 | 21.92 | 23.47 | 17.22 | 1.52 | 7.16 | 7.29 | 42.52 | **45.55** | 44.54 | 3.03 |
| | Clean | 65.51 | 36.71 | 36.97 | 48.04 | **64.62** | 62.66 | 58.22 | 63.67 | 63.14 | -2.37 | -1.48 |
| FGVCAircraft | Robust | 0.00 | 2.52 | 2.19 | 1.32 | 0.00 | 1.27 | 1.26 | 14.86 | **16.44** | 16.44 | 1.58 |
| | Clean | 20.19 | 5.43 | 5.43 | 10.80 | **19.25** | 15.88 | 16.36 | 17.16 | 17.81 | -2.38 | -1.44 |
| StanfordCars | Robust | 0.02 | 8.74 | 11.55 | 6.82 | 0.16 | 4.40 | 2.71 | 29.06 | **37.79** | 37.77 | 8.73 |
| | Clean | 51.95 | 20.91 | 25.36 | 38.68 | **52.14** | 36.21 | 44.28 | 45.29 | 46.80 | -5.15 | -5.34 |
| SUN397 | Robust | 1.25 | 19.39 | 22.58 | 14.91 | 1.72 | 8.05 | 6.40 | 42.52 | **45.83** | 44.58 | 3.31 |
| | Clean | 58.50 | 36.69 | 37.98 | 52.42 | **59.69** | 56.00 | 53.17 | 55.13 | 55.98 | -2.52 | -3.71 |
| Country211 | Robust | 0.04 | 1.79 | 2.11 | 0.85 | 0.06 | 0.67 | 0.47 | 7.42 | **8.58** | 8.54 | 1.16 |
| | Clean | 15.23 | 4.75 | 4.64 | 9.25 | **14.80** | 11.58 | 11.72 | 12.60 | 13.19 | -2.04 | -1.61 |
| Food101 | Robust | 0.66 | 13.86 | 18.57 | 11.66 | 1.20 | 13.12 | 8.03 | 55.17 | **62.00** | 61.34 | 6.83 |
| | Clean | 83.89 | 30.00 | 36.62 | 55.24 | **83.44** | 75.81 | 80.30 | 80.97 | 81.06 | -2.83 | -2.38 |
| EuroSAT | Robust | 0.03 | 11.95 | 12.51 | 10.71 | 0.15 | 2.15 | 4.57 | 12.16 | **20.19** | 20.16 | 8.03 |
| | Clean | 42.55 | 16.36 | 18.14 | 21.10 | 53.24 | 36.78 | 39.08 | 53.09 | **53.51** | 10.96 | 0.27 |
| DTD | Robust | 2.87 | 17.50 | 14.95 | 15.69 | 3.71 | 5.62 | 11.63 | 30.10 | **31.17** | 28.30 | 1.07 |
| | Clean | 40.59 | 25.16 | 21.76 | 31.97 | 37.96 | **38.92** | 34.89 | 37.18 | 38.57 | -2.02 | -0.35 |
| PCAM | Robust | 0.09 | 48.34 | 46.46 | 16.54 | 0.41 | 4.97 | 44.74 | **65.06** | 62.44 | 62.35 | -2.62 |
| | Clean | 52.62 | 49.96 | 50.04 | 52.53 | 52.73 | 52.49 | 50.38 | 53.11 | **53.46** | 0.84 | 0.35 |
| Average | Robust | 2.73 | 26.54 | 28.77 | 20.03 | 3.86 | 12.01 | 13.81 | 40.08 | **46.04** | 43.31 | 5.96 |
| | Clean | 61.51 | 40.23 | 42.26 | 50.96 | **61.61** | 57.35 | 56.62 | 58.78 | 59.60 | -1.92 | -2.01 |

Table 1: Clean and robust accuracy under the PGD-10 with $\epsilon_{\text{atk}} = 1/255$ and $\alpha_{\text{atk}} = 1/255$. Adversarial fine-tuning methods are included as baselines and fine-tuned on the Tiny ImageNet. The superscripts for adversarial fine-tuning methods indicate the attack budget used to generate adversarial images during fine-tuning. $\Delta_o$ denotes the improvement over the original CLIP model, and $\Delta_\uparrow$ indicates the gain over the previous best performance. The best performance is highlighted in bold.

without degrading clean accuracy.

**Effect of Threshold $\tau$** We investigate the impact of the discrimination threshold $\tau$ in the directional sensitivity score, which controls the counterattack strength. The corresponding results are presented in Fig. 4(d). Specifically, the discrimination threshold $\tau$ governs the trade-off between clean accuracy and robustness. When $\tau \leq 0.15$, the model achieves higher clean accuracy but lower robustness, as insufficient counterattack strength may fail to neutralize adversarial perturbations. As $\tau$ increases, robustness improves and reaches a favorable balance at $\tau = 0.17$. Further increases in $\tau$ may lead to a decline in clean accuracy due to overly aggressive counterattacks.

## Visualization Results

To further investigate the effectiveness of our method across diverse datasets, we present additional t-SNE visualizations and quantitative comparisons of counterattack diversity with robustness on twelve more datasets in the supplementary material. The results presented in Fig. 5 consistently show that counterattacks generated by DOC exhibit higher dispersion than those generated by TTC, suggesting a broader exploration of the perturbation space. Quantitatively, DOC achieves lower MeanCos scores, indicating improved diver-

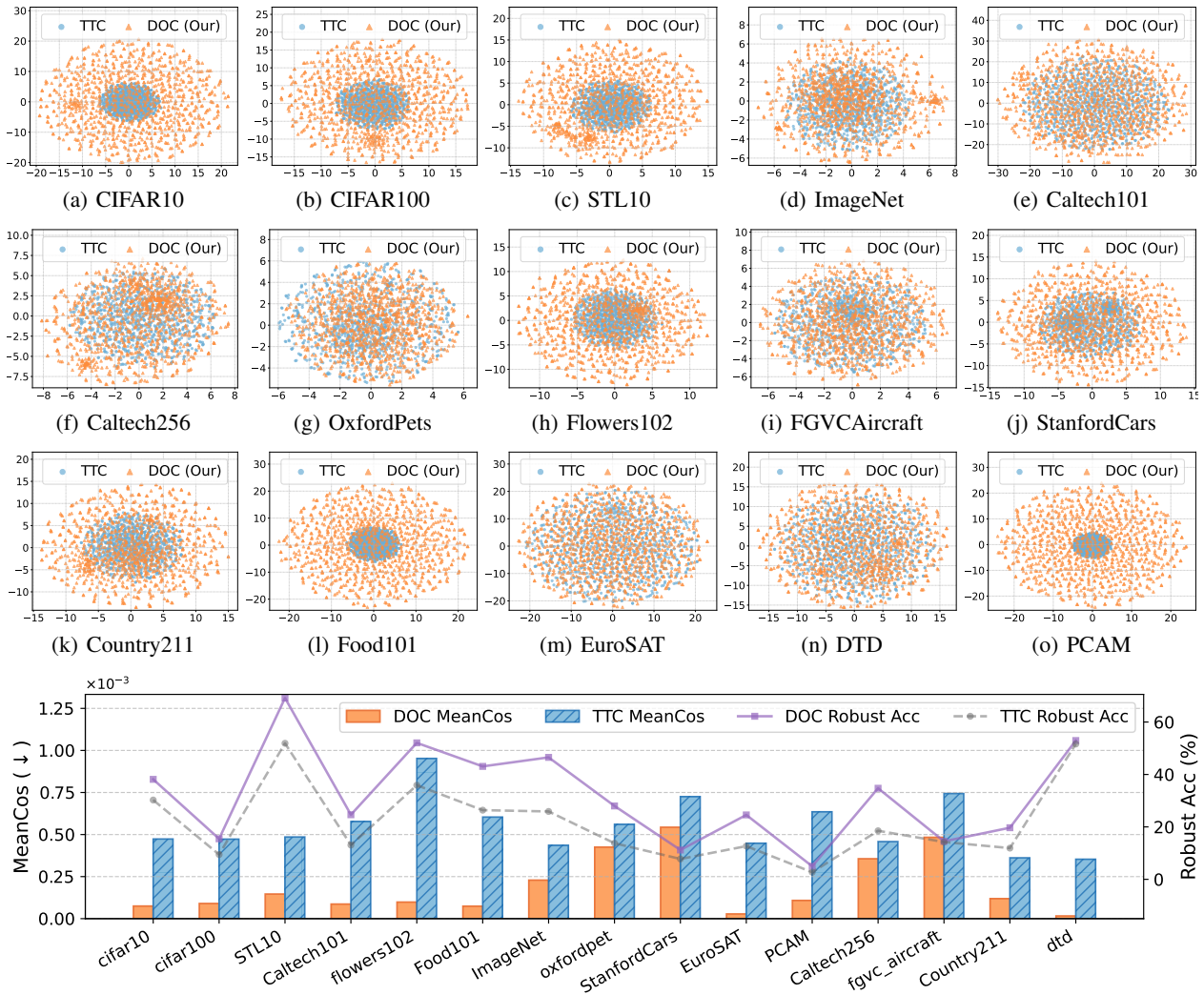

Figure 5: (a)-(c) t-SNE visualizations of counterattacks generated by DOC show greater dispersion than those from TTC, indicating improved diversity in perturbation generation. (Bottom) Comparison of mean cosine similarity (MeanCos, where lower values indicate higher diversity (Schwinn et al. 2022)) of counterattack perturbations and robust accuracy (evaluated with PGD-10, $\epsilon_{\text{atk}} = 4/255$) between DOC and TTC across seven datasets. DOC consistently achieves lower MeanCos and higher robustness, demonstrating that increased counterattack diversity significantly improves the adversarial robustness of CLIP.

sity of generated perturbations. Moreover, DOC leads to higher robust accuracy under PGD-10 evaluation with a perturbation budget of $\epsilon = 4/255$. These findings confirm that the increased diversity brought by DOC is not limited to a few datasets, but generalizes well across various scenarios.


\end{document}